\documentclass{article}
\usepackage{camera_ready_style}
\usepackage{natbib}
\usepackage{graphicx}
\usepackage{booktabs}   
\usepackage[hidelinks]{hyperref}
\usepackage{notoccite}
\usepackage{url}
\usepackage{tikz}
\usepackage{graphicx}
\usepackage{wrapfig}
\usepackage{xcolor}
\definecolor{up}{rgb}{0,0.6,.34}
\definecolor{down}{rgb}{0.9,0.2,.2}
\definecolor{br}{rgb}{0.48,0.15,.03}
\usepackage{bibunits}
\usepackage{multirow}
\usepackage{float}
\usepackage{caption, subcaption}
\usepackage{amsmath}

\usepackage{xcolor}
\usepackage{placeins}
\definecolor{dark-red}{rgb}{0.4,0.15,0.15}
\definecolor{dark-blue}{rgb}{0.15,0.15,0.4}
\definecolor{medium-blue}{rgb}{0,0,0.5}
\hypersetup{
  colorlinks, linkcolor={dark-blue},
  citecolor={dark-blue}, urlcolor={medium-blue}
}

\usepackage{comment}

\usepackage{algpseudocode,algorithm,algorithmicx}

\usepackage{times}

\title{Greedy Policy Search:\\A Simple Baseline for Learnable Test-Time Augmentation}

\author{ {\bf Dmitry Molchanov$^*$~$^{1,2}$~~~~Alexander Lyzhov$^*$~$^{1,3,4}$~~~~Yuliya Molchanova$^*$~$^1$~~~~Arsenii Ashukha$^*$~$^{1,2}$~~~~Dmitry Vetrov~$^{2,1}$} \\
\\
$^{1}$Samsung AI Center Moscow\\
$^{2}$Samsung-HSE Laboratory, National Research University Higher School of Economics\\~~$^{3}$National Research University Higher School of Economics\\~~$^{4}$Skolkovo Institute of Science and Technology\\
\\
\\
}

\usepackage{printlen}

\begin{document}

\maketitle


\ \\ 

\begin{abstract}

Test-time data augmentation---averaging the predictions of a machine learning model across multiple augmented samples of data---is a widely used technique that improves the predictive performance.
While many advanced learnable data augmentation techniques have emerged in recent years, they are focused on the training phase.
Such techniques are not necessarily optimal for test-time augmentation and can be outperformed by a policy consisting of simple crops and flips.
The primary goal of this paper is to demonstrate that test-time augmentation policies can be successfully learned too.
We~introduce \emph{greedy policy search} (GPS), a simple but high-performing method for learning a policy of test-time augmentation.
We demonstrate that augmentation policies learned with GPS achieve superior predictive performance on image classification problems, provide better in-domain uncertainty estimation, and improve the robustness to domain shift.
\end{abstract}


{\let\thefootnote\relax\footnote{$^*$ Equal contribution}}


\section{INTRODUCTION}

Convolutional neural networks (CNNs) have become a \emph{de facto} standard for problems with complex data that contain a lot of label-preserving symmetries.
Such architectures use spatially invariant operations that have been specifically designed to reflect the symmetries present in data.
These architectural choices are not enough, so data augmentation that artificially expands a dataset with label-preserving transformations is used during training to further promote the invariance to such symmetries.

\begin{figure}[!t]
\centering
\vspace{0.85cm}
\includegraphics[width=0.45\textwidth]{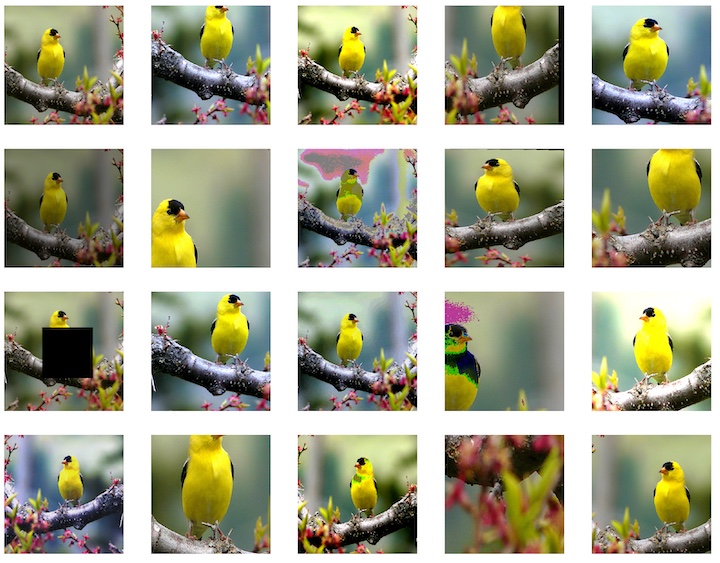}
\caption{
A sample from the test-time data augmentation policy learned by greedy policy search for EfficientNet-B5 on ImageNet.
Averaging the predictions across samples from the policy outperforms the conventional multi-crop evaluation by a wide margin.
}
\label{fig:first_vis}
\end{figure}

Training with data augmentation has been used for a long time to improve the predictive performance of machine learning and pattern recognition algorithms \citep{yaeger1997effective,simard2003best, krizhevsky2012imagenet}.
Earlier techniques enlarge datasets with a handcrafted set of transformations, such as scale, translation, rotation, and require manual tuning of augmentation strategies. 
Recent works explore learnable and more diverse strategies of data augmentation \citep{cubuk2019autoaugment, cubuk2019randaugment, lim2019fast}.
These strategies have become a standard component of training powerful deep learning models \citep{tan2019efficientnet}.


Even when learning with data augmentation, CNNs are still not perfectly invariant to all the symmetries present in the data distribution.
Therefore, test-time augmentation---averaging the predictions of a model across multiple augmentations of an object---often increases predictive performance.
A special case of test-time augmentation called multi-crop evaluation has even become a standard evaluation protocol for large scale image classification \citep{krizhevsky2009learning, simonyan2014very, he2016deep}. 
Test-time augmentation is, however, limited to simple transformations and usually does not benefit from using a more diverse augmentation policy, e.g.\ the one used during training.


In this work, we aim to demonstrate that test-time augmentation of images can benefit more from a wide range of diverse data augmentations if their composition is learned.
We introduce \emph{greedy policy search} (GPS), a simple algorithm that learns a policy for test-time data augmentation based on the predictive performance on a validation set.
In an ablation study, we show that optimizing the calibrated log-likelihood \citep{Ashukha2020Pitfalls} is a crucial part of the policy search algorithm, while the default objectives---accuracy and log-likelihood---lead to a significant drop in the final performance.

Our evaluation is performed on the following problems: conventional image classification, in-domain uncertainty estimation, and classification under dataset shift.
We demonstrate that test-time augmentation policies found by GPS (see an example on Figure~\ref{fig:first_vis}) outperform other data augmentation baselines significantly on a wide range of deep learning architectures from VGG-style networks \citep{simonyan2014very} to the recently proposed EfficientNets \citep{tan2019efficientnet}.
GPS provides consistent improvements in the performance of ensembles, models trained with powerful train-time data augmentation techniques such as AutoAugment \citep{cubuk2019autoaugment} and RandAugment \citep{cubuk2019randaugment}, as well as models trained without advanced data augmentation.
We also show that the obtained policies transfer well across different architectures.

\section{RELATED WORK}

\paragraph{Test-time augmentation}
The test-time data augmentation (TTA) has been present for a long time in deep learning research. 
\cite{krizhevsky2012imagenet} averaged the predictions of an image classification model over random crops and flips of test data.
This became a standard evaluation protocol \citep{krizhevsky2009learning, simonyan2014very, he2016deep}.
\cite{shorten2019survey} provided an extensive survey of data augmentation for deep learning including test-time augmentation, pointing out several successful applications of TTA in medical imaging.
As one example, \cite{wang2019aleatoric} show that TTA improves uncertainty estimation for medical image segmentation.
\cite{pang2019mixup} demonstrated that mixup data augmentation \citep{zhang2017mixup} can be applied during testing, improving defense against adversarial attacks on image classifiers.

\paragraph{Learnable train-time augmentation}
Data augmentation is more commonly applied during training rather than during inference. 
Seeking to improve train-time augmentation, a recent line of works starting from \cite{cubuk2019autoaugment} explored the practice of adapting it to peculiarities of a specific dataset.
AutoAugment \citep{cubuk2019autoaugment} learns an augmentation policy with reinforcement learning and requires a repetition of an expensive model training for each iteration of the policy search algorithm.
Subsequent works proposed more efficient methods of policy search for training set augmentation \citep{ho2019population, cubuk2019randaugment, lim2019fast, zhang2019adversarial}.

\paragraph{Ensembling}
Neural network ensembling---computing predictions using a distribution over neural networks instead of a single model---improves performance on various machine learning problems. Often, ensembling involves obtaining a set of trained neural networks and averaging their predictions on each test object. There are many methods of ensembling \citep{srivastava2014dropout,
blundell2015weight,
lakshminarayanan2017simple, huang2017snapshot}, differing in time and memory requirements, diversity of ensemble members and performance.

\paragraph{Sub-ensemble selection}
Even though a single model is used for TTA, it makes sense to see the TTA as an ensemble of different models, each with its own augmentation sub-policy.
The specific members in this ensemble can be selected from a variety of discrete possibilities.
Historically, ensemble pruning methods have been applied for such optimization problems.
\cite{partridge1996engineering} introduced a heuristic which can serve as a rule of selection of ensemble members.
\cite{fan2002pruning,caruana2004ensemble} described and used another, simpler, greedy ensemble pruning method which is the one that we adopt in this work for test-time augmentation.

\section{LEARNABLE TEST-TIME AUGMENTATION}


In this section we discuss the training of test-time augmentation policy for image classification problems.


\textbf{Policy~}
We define a test-time augmentation (TTA) policy $P$ as a set of sub-policies $\{s_i(\cdot)\}$.
A sub-policy $s(\cdot)$ consists of $N_{s}$ consecutively applied
image transformations $t_j(\cdot, M_j)$, $j~\in~\{1, \dots, N_{s}\}$, where $t_j$ 
is one of the predefined image operations, $M_j\geq 0$ being its magnitude.
The transformations that we use and their respective typical magnitudes are listed in Appendix~\ref{sec:training_details}.
A visualization of these transforms is presented in Figure~\ref{fig:ra}.

\textbf{Inference~} During inference, the predictions are averaged across samples of different sub-policies:
\begin{equation}
\label{eq:tta}
    \pi_\theta^{P}(x)=\frac{1}{|P|}\sum_{s\in P}p(y\,|\,s(x),\theta).
\end{equation}




\subsection{Naive approaches to test-time augmentation}

Common test-time augmentation policies consist of sub-policies that are sampled independently from a fixed distribution.
For example, a single sub-policy may consist of randomly resized crops and horizontal flips.
A potential alternative is to use the same policy that has been learned for training (e.g.\ a policy obtained with RandAugment \citep{cubuk2019randaugment} or AutoAugment \citep{cubuk2019autoaugment}) to perform test-time augmentation.
A possible motivation behind this choice is that such a policy might reflect the specifics of a particular dataset or architecture better.


For simplicity, we use a slightly modified set of PIL transforms that is commonly used for learning the training time augmentation policies as test-time augmentation transformation options.

\begin{figure}[t]
\centering
\includegraphics[width=.75\columnwidth]{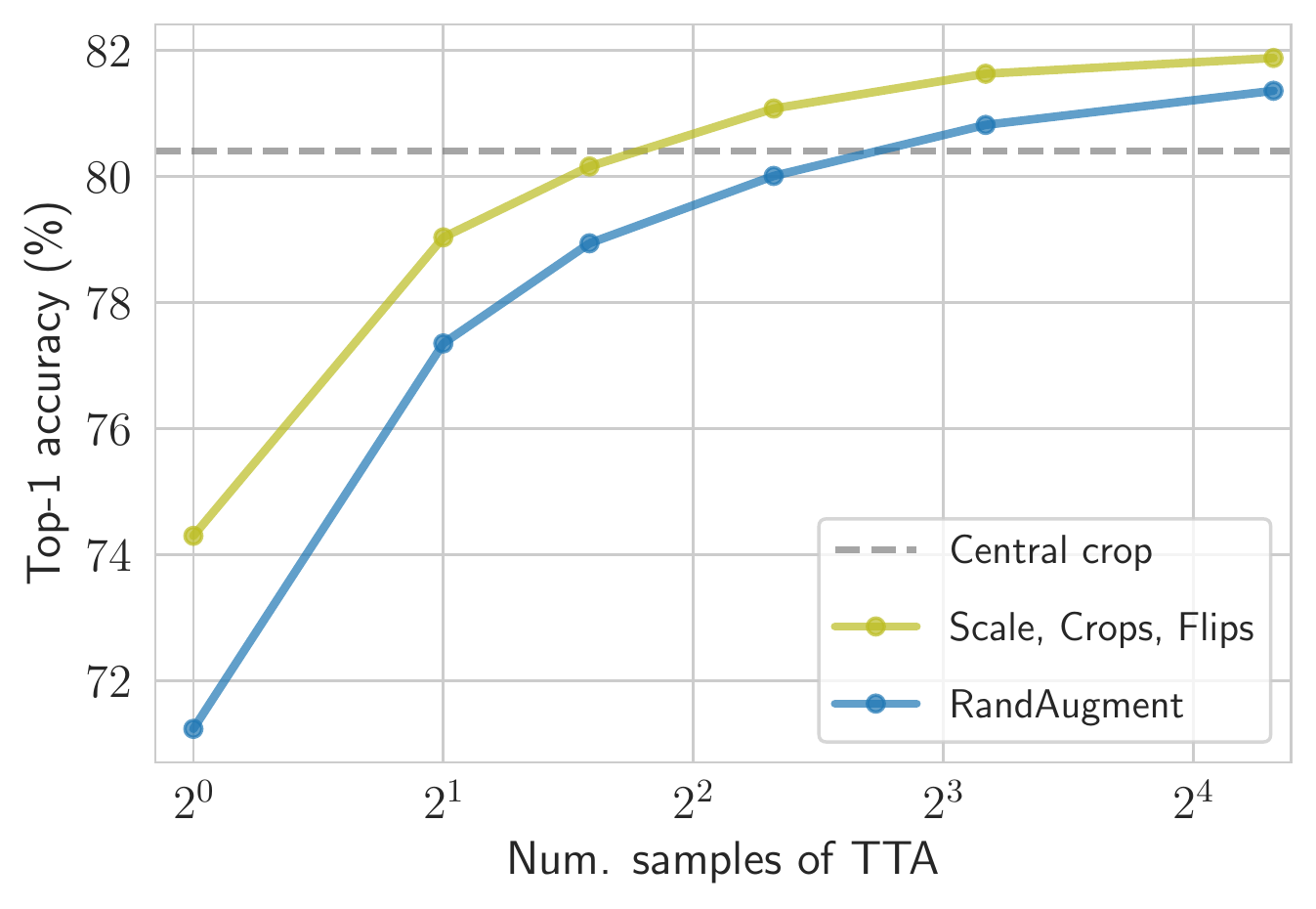}
\caption{
    Accuracy of EfficientNet B2 (trained with \mbox{RandAugment}) on ImageNet for two TTA strategies: scale-crop-flip augmentation, and RandAugment (the same as during training). 
    The scale-crop-flip policy outperforms the RandAugment policy and the effect still holds for large number of samples.
    This example demonstrates that the policy learned for training is not necessarily optimal for test-time augmentation.
    }
\label{fig:train_vs_clean}
\end{figure}

Our experiments indicate that in some cases (Figure \ref{fig:train_vs_clean}) a TTA policy that was learned for training performs worse than the default policy consisting of random scalings, crops and flips.
This means that the process of learning a policy for training does not necessarily result in a good TTA policy.
A natural alternative is to learn the TTA policy for a trained neural network by directly optimizing some TTA performance objective.
For example, we can parameterize a policy with a magnitude parameter shared across all transformations, as in RandAugment \citep{cubuk2019randaugment}, and find the optimal magnitude using grid search.
As we show in Figure~\ref{fig:mgrid}, the optimal magnitude for test-time augmentation is different from the optimal magnitude for training.
To push the idea of direct optimization of TTA performance further, we employ the greedy ensemble pruning for TTA.
The resulting method, \emph{greedy policy search}, can be considered a simple yet strong baseline for more advanced discrete optimization method like reinforcement learning, used in AutoAugment \citep{cubuk2019autoaugment}, or Bayesian optimization, used in Fast AutoAugment \citep{lim2019fast}.

\subsection{Greedy policy search}

We introduce \emph{greedy policy search} (GPS) as a means of demonstrating that learnable policy for test-time augmentation can boost the predictive performance, uncertainty estimates and robustness of deep learning models.

\textbf{Greedy policy search~} 
GPS starts with an empty policy and builds it in an iterative fashion.
It searches for the sub-policy that provides the largest performance gain when added to the current policy.
This selection step is repeated until a policy of the desired length is built.
To make the procedure computationally efficient, we first draw a pool of candidate sub-policies from a prior distribution over sub-policies $p(s)$.
We precompute the predictions on all these sub-policies so that the sub-policy selection step could be performed in the space of predictions without passes through the neural network.
Both the pool generation and the selection procedure are embarrassingly parallel, so the resulting algorithm is efficient and easily scalable.
The whole procedure is summarized in Algorithm~\ref{alg:gps}.

\begin{figure*}[t!]
\centering
\includegraphics[width=0.85\textwidth]{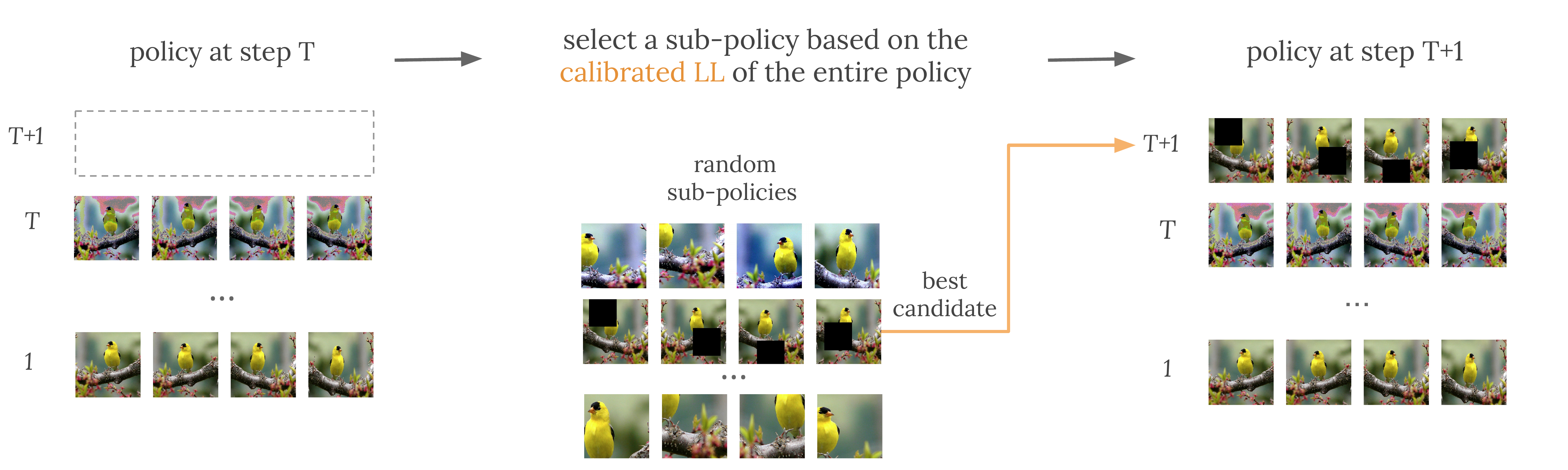}
\caption{
An illustration of one step of the greedy policy search algorithm.
Each step selects a sub-policy that provides the largest improvement in calibrated log-likelihood of ensemble predictions and add it to the current policy.
}
\label{fig:scheme}
\end{figure*}
\textbf{Optimization criterion~}
The criteria of predictive performance that are often used as objectives for policy, architecture or hyperparameter search are classification accuracy and log-likelihood.
We find, however, that these criteria are ill-suited for TTA policy search.
As we discuss in Section~\ref{subsec:cll_section}, the log-likelihood is unable to fairly judge the performance of test-time augmentation, and the accuracy is typically too noisy to provide an adequate signal for learning a well-performing TTA policy.
We follow \cite{Ashukha2020Pitfalls} and use the calibrated log-likelihood instead.
The calibrated log-likelihood is defined as the log-likelihood measured after the post-hoc temperature scaling \citep{guo2017calibration}.
The temperature scaling is typically performed by optimizing the validation log-likelihood w.r.t.\ the temperature $\tau$ of the $\mathrm{softmax(\cdot / \tau)}$ function used to obtain the predictions.
Our experiments show that the calibrated log-likelihood is the key ingredient of GPS.
This objective is suited for learning TTA policies better than both accuracy and conventional uncalibrated log-likelihood.

\begin{algorithm}[t!]
  \caption{\label{alg:gps} Greedy Policy Search (GPS)}
  \begin{algorithmic}  
    \Require{Trained neural network $p(y\,|\,x, \theta)$}
    \Require{Validation data $X_{val}, y_{val}$}
    \Require{Pool size $B$, policy size $T$}
    \Require{Prior over sub-policies $p(s)$}
    \State $S \gets \emptyset$ \Comment{Pool of candidate sub-policies}
    \For{$i \gets 1$ to $B$}
        \State $s_i \sim p(s)$
        \State $S \gets S \cup \{s_i\}$\Comment{Add $s_i$ to pool}
        \State $\pi_{val}^{s_i} \gets p(y\,|\,s_i(X_{val}), \theta)$
        \Comment{Predict with $s_i$}
    \EndFor
    \State $P\gets \emptyset$ \Comment{GPS policy}
    \State $\pi_{val}^{P} \gets 0$
    \Comment{Predictions made with GPS policy}
    \For{$t \gets 1$ to $T$}
    \\\Comment{Choose the best sub-policy $s^*$ based on \textbf{calibrated}\\\ \ \ \ \ \,\textbf{log-likelihood} on validation:}
        \State $s^* \gets {\displaystyle\arg\max_{s\in S}}\ \mathrm{cLL}\left( \frac{t-1}{t}\pi_{val}^P + \frac{1}{t}\pi_{val}^{s};\ \ y_{val}\right)$
        \State $\pi_{val}^{P} \gets \frac{t-1}{t}\pi_{val}^P + \frac{1}{t}\pi_{val}^{s^*}$
        \Comment{Update predictions}
        \State $P\gets P \cup \{s^*\}$ \Comment{Update policy}
    \EndFor\\
    \Return policy P
  \end{algorithmic}  
\end{algorithm}

\begin{figure*}[t]
\centering
\includegraphics[width=1\textwidth]{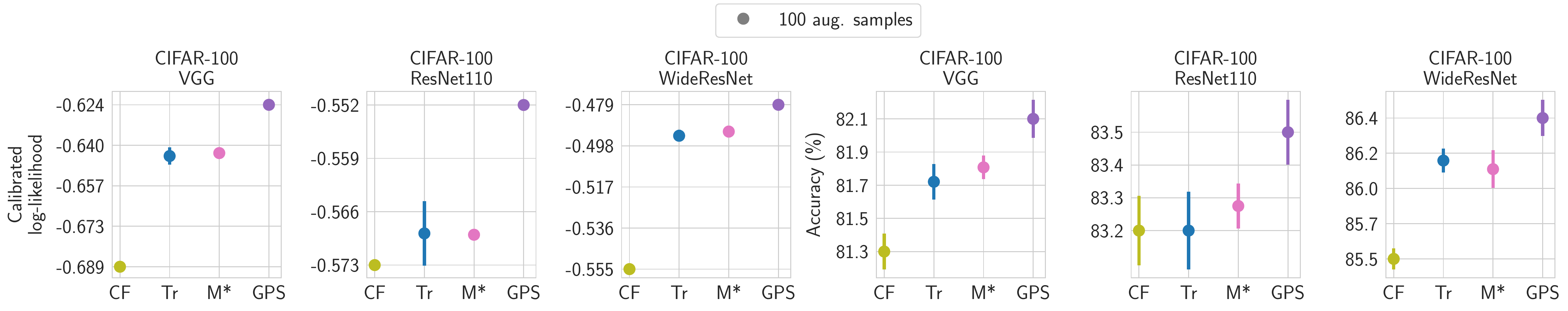}
\includegraphics[width=1\textwidth]{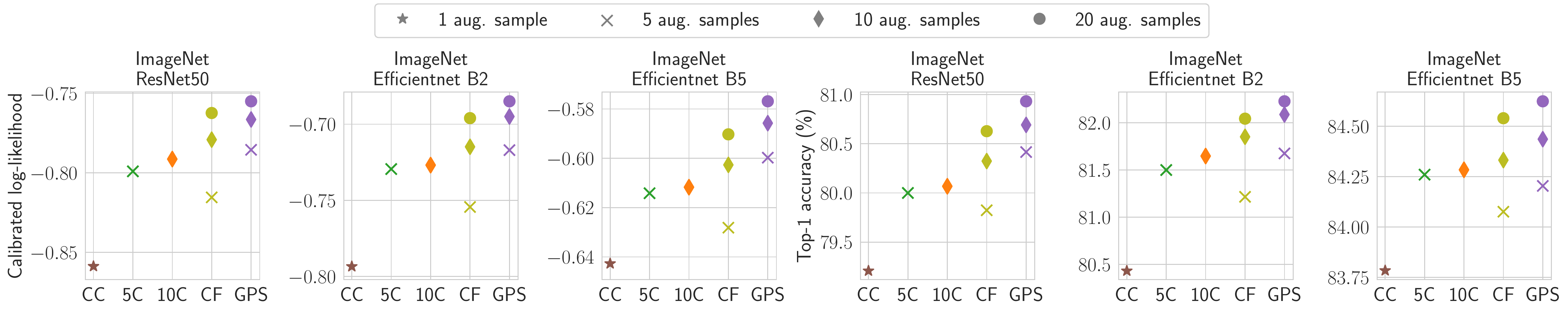} 
\caption{
Performance of various test-time augmentation strategies on clean test set of CIFAR-100 dataset (top) and ImageNet (bottom).
CC: central crop.
CF: random crops and horizontal flips.
Tr: augmentation used for training (modified RandAugment with $M=45$).
$M^*$: modified RandAugment with $M$ found by grid search.
5/10C: 5/10-crop evaluation (four corner crops, one center crop for 5C; five crops with horizontal flips for 10C).
Greedy policy search (GPS) consistently outperforms all other methods in both the calibrated log-likelihood and accuracy.
The results for CIFAR-100 have been averaged over five runs of TTA.
}
\label{fig:in-domain-cifar100}
\end{figure*}

\section{EXPERIMENTS} \label{sec:exps}
We perform experiments with greedy policy search on a variety of architectures on CIFAR-10/100 and ImageNet classification problems.
On CIFAR-10/100 datasets \citep{krizhevsky2009learning}, we use VGG16 \citep{simonyan2014very}, PreResNet110 \citep{he2016deep} and WideResNet28x10 \citep{zagoruyko2016wide}.
On ImageNet \citep{russakovsky2015imagenet}, we use ResNet50 and EfficientNet B2/B5/L2 \citep{tan2019efficientnet}.
PyTorch \citep{paszke2017automatic} is used for all experiments.
The source code is available at \url{https://github.com/bayesgroup/gps-augment}.

\textbf{Training~} CIFAR models were trained for 2000 epochs using a modified version of RandAugment with $N=3$ transformations for each image, where the magnitude of each transformation for each image has been drawn from the uniform distribution $\widetilde{M}\sim\mathcal{U}\left[0, 45\right]$.
We provide the details of training these models in Appendix~\ref{sec:training_details}.

We reused the publicly available snapshots\footnote{\url{https://github.com/tensorflow/tpu/tree/master/models/official/efficientnet}}$^{\tiny{\&}}$\footnote{\url{https://github.com/rwightman/pytorch-image-models}} of ImageNet models. 
EfficientNets B2/B5 were trained with vanilla RandAugment, 
EfficientNet L2 was trained with Noisy Student~\citep{xie2020self} and RandAugment, 
ResNet50 was trained with AugMix~\citep{hendrycks*2020augmix} and RandAugment.

\textbf{Policy search~} To obtain the results on CIFAR datasets, we first train all our models with the same stratified train-validation split (we use 45000 objects for training and 5000 objects for validation), and perform GPS or magnitude grid search on the validation set.
We then retrain all models on the full training set, and evaluate them with the obtained policies.
Since we did not train the ImageNet models, we split the validation set in half with a stratified split, use the first half for policy search and report the results for the second half.
We use approximately 1000 sub-policies in the candidate pools for GPS, and describe the construction of the pools in Appendix~\ref{sec:training_details}.

\textbf{Evaluation~} Following \cite{Ashukha2020Pitfalls}, we use the calibrated log-likelihood as our main evaluation metric for in-domain uncertainty estimation, and we reuse their ``test-time cross-validation'' procedure to perform calibration.
The test set is divided in half, the optimal temperature is found on the first split, and the metrics are evaluated on the second split.
We average the metrics across five random splits.
While it is possible to optimize the temperature on a validation set, we stick with test-time cross-validation for convenience since the optimal temperature is different for each TTA policy and for each number of samples for TTA (see Figure~\ref{fig:temp} for details).
The optimal temperature has a very low variance, and the values found on the validation set closely match the values found during test-time cross-validation.


\subsection{In-domain predictive performance}


Greedy policy search achieves better predictive performance compared to all of the following: conventional test-time augmentation techniques (e.g.\ random crops and flips), reuse of policy learned during training, and a more advanced baseline (RandAugment with magnitude grid search).
The results for CIFAR-100 and ImageNet are presented in Figure~\ref{fig:in-domain-cifar100}, and the results for CIFAR-10 are presented in Figure~\ref{fig:in-domain-cifar10}, numerical results can be found in Tables~\ref{table:acc-nll-ImageNet},~\ref{table:acc-nll-cifars}.


When using the same amount of samples, GPS has the same test-time computational complexity as vanilla test-time augmentation or the standard multi-crop evaluation, yet achieves a better predictive performance. 
Once the GPS policy is found or transferred from a different model or dataset, the gain in the predictive performance can be obtained for free.

Aside from test-time data augmentation, there are other techniques that allow one to use ensembling during test time with almost no training overhead.
Such methods as variational inference \citep{blundell2015weight}, dropout \citep{srivastava2014dropout}, K-FAC Laplace approximation
are praised as ways to hide an ensemble inside a single model using a stochastic computation graph.
It was, however, recently shown that these techniques are typically significantly outperformed by test-time augmentation with random crops and flips \citep{Ashukha2020Pitfalls} in conventional image classification benchmarks (CIFAR and ImageNet classification).
Since GPS outperforms vanilla TTA, it outperforms these techniques as well.
However, GPS can be combined with ensembling techniques to further improve their performance (see Section~\ref{subsec:ens}).

\subsection{What metric to use for policy search?} \label{subsec:cll_section}
Any policy search procedure that relies on optimizing the validation performance requires a metric to optimize.
Common predictive performance metrics are classification accuracy and log-likelihood.

\begin{table}[]
\centering
\resizebox{\columnwidth}{!}{\begin{tabular}{rcccc}
\toprule
\multicolumn{1}{l}{}  & \begin{tabular}[c]{@{}c@{}}GPS\\criterion\end{tabular}         & \multicolumn{1}{c}{VGG} & \multicolumn{1}{c}{ResNet110} & \multicolumn{1}{c}{WideResNet} \\
\midrule
\multirow{3}{*}{\rotatebox[origin=c]{90}{Acc.(\%)}} 
     & Acc.  & $81.17\pm0.15$  & $83.01\pm0.18$  & $85.71\pm0.10$                  \\
     & LL            & $81.89\pm0.07$  & $\mathbf{83.55\pm0.09}$  & $86.22\pm0.05$  \\
     & cLL & $\mathbf{82.21\pm0.17}$  & $\mathbf{83.54\pm0.06}$  & $\mathbf{86.44\pm0.05}$  \\
\midrule
\multirow{3}{*}{\rotatebox[origin=c]{90}{cLL}}
     & Acc.  & $-0.837\pm0.003$  & $-0.691\pm0.001$  & $-0.661\pm0.003$                  \\
     & LL            & $-0.640\pm0.001$  & $-0.560\pm0.001$  & $-0.489\pm0.001$  \\
     & cLL & $\mathbf{-0.623\pm0.001}$  & $\mathbf{-0.552\pm0.001}$  & $\mathbf{-0.479\pm0.001}$ 
\\
\bottomrule
\end{tabular}}
\caption{Performance of greedy policy search using different metrics as a search objective, measured on CIFAR-100 dataset. Calibrated log-likelihood results in superior performance across all tasks and metrics. {The results have been averaged over five runs of TTA.}}
\label{table:gps_metrics}
\end{table}

\begin{figure*}[t]
\includegraphics[width=\textwidth]{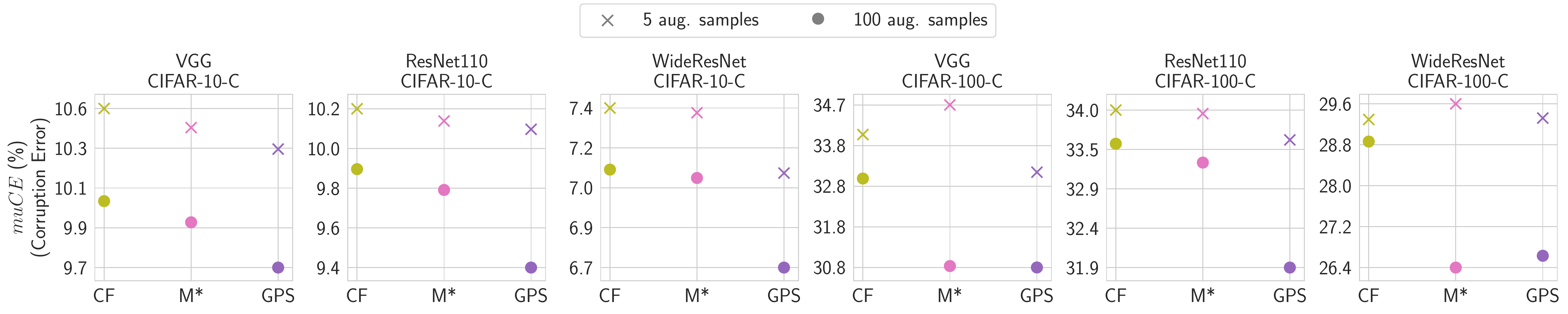}
\caption{
{
Mean unnormalized corruption error {($muCE$)} on corrupted versions of CIFAR datasets for various test-time augmentation strategies: random crops and horizontal flips (CF), modified RandAugment with $M$ found by grid search~($M^*$) and GPS policy (GPS).
Learnable TTA methods are run on clean, uncorrupted data.
In most cases, GPS policies are more robust to the domain shift compared to alternatives.
}
}
\label{fig:ood-cifar}
\end{figure*}
\begin{figure}[t]
\includegraphics[width=\columnwidth]{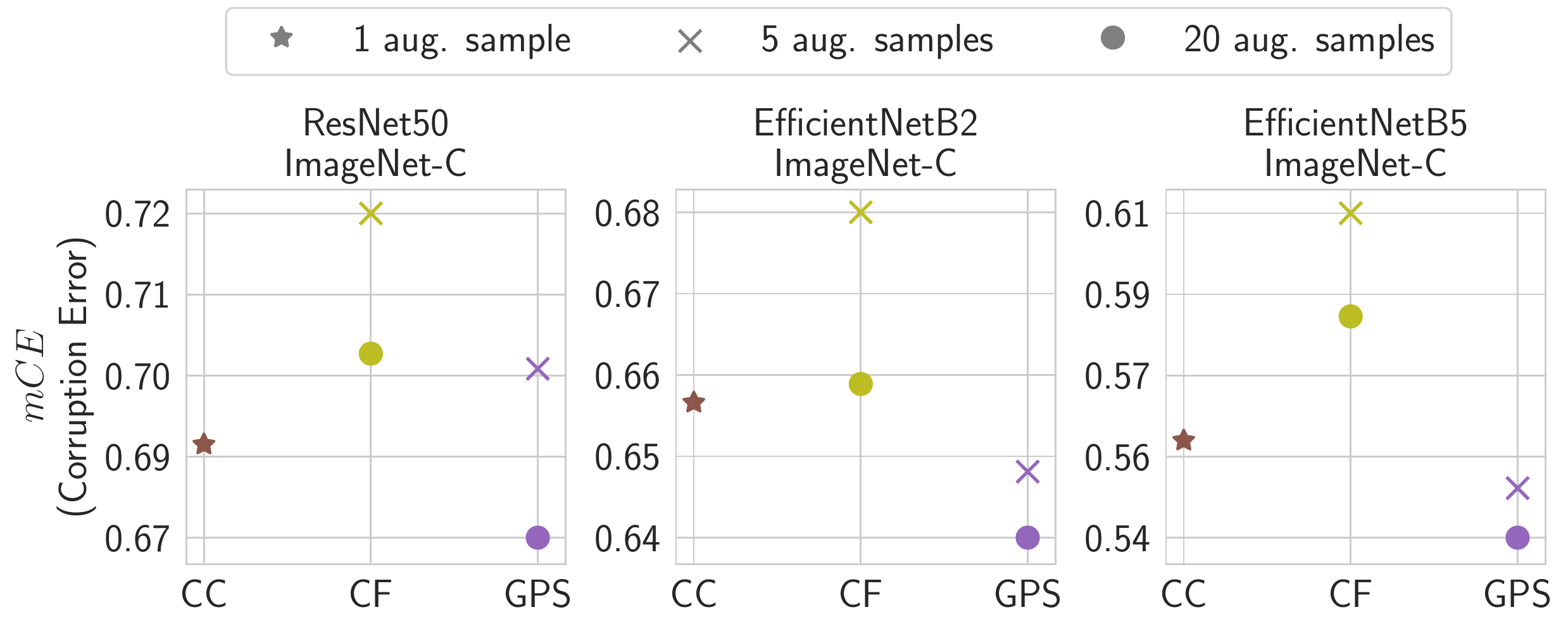}
\caption{
{Mean corruption error {($mCE$)} on ImageNet-C for various test-time augmentation strategies: central crop (CC), random scale-crop-flip transformation (CF), GPS policy trained on the clean data (GPS).
GPS  policy  outperforms  non-learnable test-time augmentation strategies under domain shift. 
 }
}
\label{fig:ood-imagenet}
\end{figure}

Both of these metrics have problems.
The plain log-likelihood cannot be used for a fair comparison of different techniques, especially in the test-time augmentation setting \citep{Ashukha2020Pitfalls}.
The authors suggest switching to calibrated log-likelihood (cLL) instead.
The problem with the log-likelihood is that it can dismiss a good model that happened to be miscalibrated, but can be fixed by temperature scaling.
With test-time augmentation it is often the case that {the optimal temperature of the predictive distribution} drastically changes with the number of samples (see Figure~\ref{fig:temp}).
The accuracy, in turn, appears to be too noisy to provide robust learning signal for greedy optimization.

To evaluate the influence of the objective function, we run GPS for a VGG, a PreResNet110 and a WideResNet28x10 on CIFAR-100 dataset.
The pool of candidate sub-policies and the resulting length of sub-policy is kept the same for all methods, as described in Section~\ref{sec:exps}.
We evaluate three different objectives for GPS: classification accuracy, log-likelihood and calibrated log-likelihood.
The results are presented in Table~\ref{table:gps_metrics}.
We find that optimizing the calibrated log-likelihood consistently outperforms other metrics in terms of both accuracy and calibrated log-likelihood.

To better see how the metrics fail, we evaluate test-time RandAugment policies with different magnitudes $M$.
As one can see from Figure~\ref{fig:mgrid}, the optimal value of $M$ is different for different metrics.
The accuracy is too noisy to reliably find the optimal $M$.
The log-likelihood provides a very conservative value of $M$ since large magnitudes decalibrate the model.
On the contrary, the calibrated log-likelihood does not suffer from this problem and results in a better value of $M$.

\subsection{Robustness to domain shift}

Despite the natural human ability to correctly recognize an object given an image with visual perturbations, neural networks are typically very sensitive to changes in the data distribution.
As for now, models suffer a significant performance loss even under a slight domain shift \citep{ovadia2019can}.
To explore how different test-time augmentation strategies influence the robustness to domain shift, we use the benchmark, proposed by \cite{hendrycks2018benchmarking}.

We perform an evaluation of TTA methods on CIFAR-10-C, CIFAR-100-C and ImageNet-C datasets with 15 corruptions $C$ from groups \emph{noise}, \emph{blur}, \emph{weather} and \emph{digital}.
These {datasets consist of the} test sets of the corresponding {original} datasets with applied corruption transforms $c \in C$ with five different severity levels $s$, $1 \leq s \leq 5$.
For a given corruption $c$ at severity level $s$ we compute the error rate $E_{c,s}$.
On CIFAR datasets for each corruption we compute the unnormalized corruption error $uCE_c = \frac{1}{5} \sum_{s=1}^{5}E_{c,s}$, as proposed by \cite{hendrycks*2020augmix}, whereas for ImageNet-C we normalize the corruption error by the central crop performance of AlexNet: $CE_c=\sum_{s=1}^5 E_{c, s} / \sum_{s=1}^5 E^{AlexNet}_{c, s}$.
We obtain the final metric {$muCE$ or $mCE$} by averaging the corruption errors ($uCE_c$ or $CE_c$) over different corruptions $c \in C$.
We report these metrics for the policies found using the clean validation data (the same policies as in other experiments), and compare our method with several baselines.
The results are presented in Figures~\ref{fig:ood-cifar}~and~\ref{fig:ood-imagenet} and in Tables~\ref{table:ce-cifarsC}~and~\ref{table:ce-ImageNetC}.

We use the same stratified validation-test split as the one we used for policy search.
It should be noted that ImageNet-C has a different data format compared with ImageNet: it consists of images with pre-applied central cropping which shrinks the resolution down to $224\times 224$.
For this experiment, we use the same magnitudes for scale and crop transforms as before for all the considered policies even though these magnitudes were set on full-resolution images.
Although such choice may not be optimal, it is consistent, and still leads to a substantial improvement over the central crop baseline.
Ideally, the ImageNet-C dataset should be modified to contain corrupted full-resolution images to establish a unified benchmark for models, designed for different resolutions and for non-standard inference techniques such as test-time data augmentation.

Even though the corruptions of ImageNet-C do slightly intersect with the augmentation transformations used during training, this does not favor GPS over other methods.

Surprisingly, policies trained on clean validation data work decently for corrupted data.
In most cases, GPS outperforms both the conventional baselines and RandAugment with the optimal (for the clean validation set) magnitude $M^*$.
Somewhat counter-intuitively, we find that extreme augmentations (see Figure~\ref{fig:ra_augs_cifars}) of data that is already corrupted leads to a significant performance boost as compared to conservative crops and flips.
{Not only does this demonstrate} the efficiency of learnable TTA, it also shows that the policy does not overfit to clean data and consists of augmentations that are useful in other settings.


Although ensembling is a popular way to mitigate dataset shift \citep{ovadia2019can}, we do not compare model ensembles with TTA in this setting.
As noted by \citep{Ashukha2020Pitfalls} and as we show in Section~\ref{subsec:ens}, ensembling and test-time augmentation are complementary practices and can be combined to boost the performance.
We expect this combination to work well in the setting of domain shift.

\begin{figure}[t!]
\includegraphics[width=0.9\columnwidth]{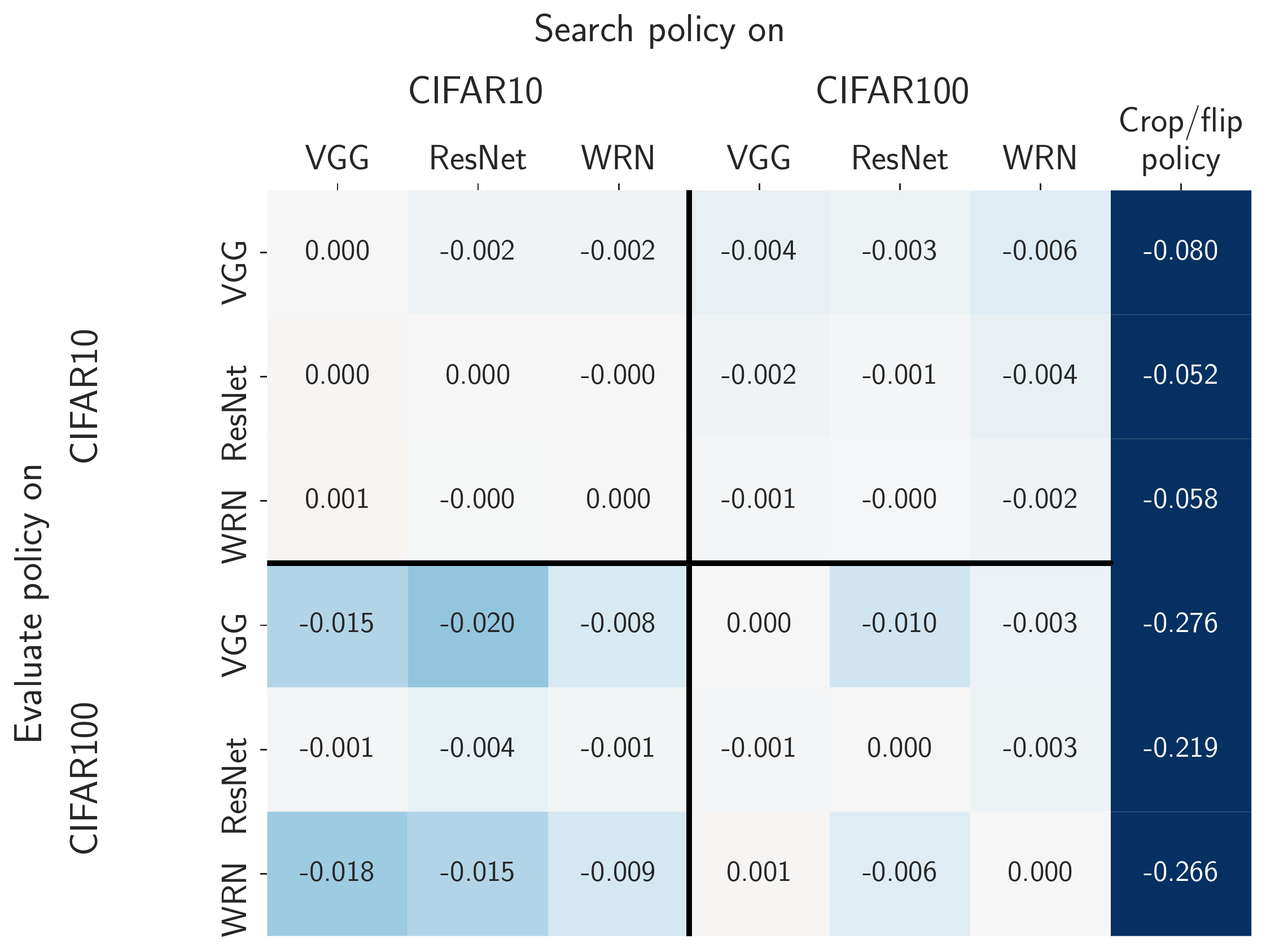}
\caption{
{The change in cLL when switching from a GPS policy learned for one dataset-architecture pair to a GPS policy learned for another dataset-architecture pair.}
Policy transfer outperforms random crops and flips in all considered cases.
Negative numbers mean that TTA works best when the policy is evaluated on the same architecture and dataset as used for policy search.
{The results have been averaged over five runs of TTA.}
}
\label{fig:transfer-cifar}
\end{figure}

\begin{figure}[t!]
\includegraphics[width=\columnwidth]{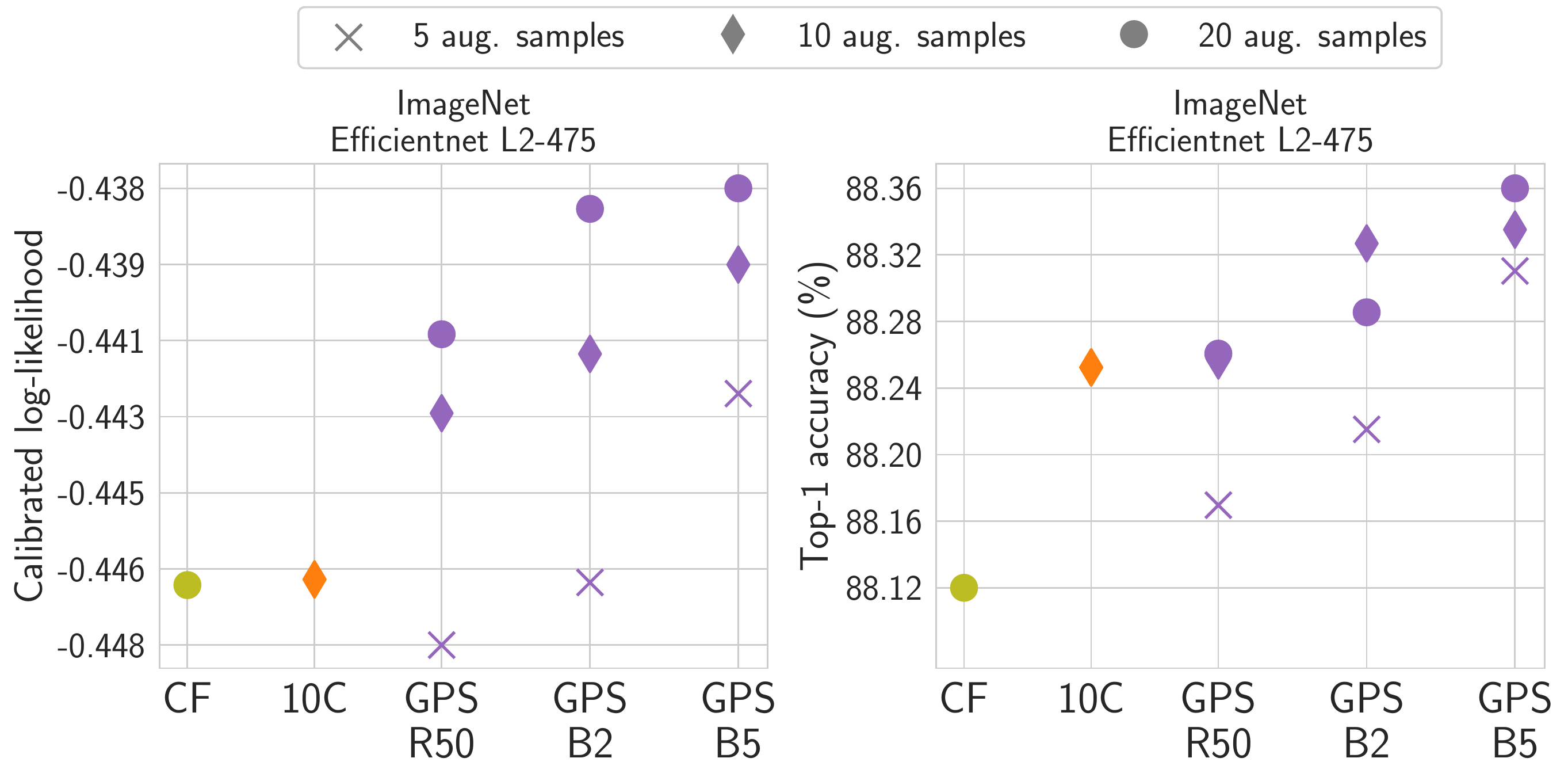}
\caption{
{
Policies learned with GPS for ResNet-50 (GPS R50), EfficientNet B2 (GPS B2), and EfficientNet B5 (GPS B5) models transfer well to the larger EfficientNet L2 architecture and outperform conventional baselines for multi-crop evaluation: random scale-crop-flip transformation (CF) and multi-crop evaluation with 5 crops and 2 horizontal flips for each crop (10C).}
}
\label{fig:transfer-imagenet}
\end{figure}

\subsection{Policy transfer}

We evaluate the policies found by GPS on other architectures and datasets in order to test their generality.
The change in calibrated log-likelihood when transferring the policies across CIFAR datasets and architectures is reported in Figure~\ref{fig:transfer-cifar}.
The decrease in performance is not dramatic, and the transferred policies still outperform standard random crop and flip augmentations.
We observe that keeping the same dataset during transfer is more important than keeping the same architecture.

We also transfer the GPS policies found on ImageNet for ResNet50, EfficientNet-B2 and EfficientNet-B5 to an even larger architecture, EfficientNet-L2, and show the results in Figure~\ref{fig:transfer-imagenet}.
We observe that all of these policies transfer to a larger architecture well, and outperform the vanilla test-time augmentation policy and multi-crop evaluation significantly.

We do not transfer policies from CIFAR to ImageNet and vice versa since the image preprocessing for these datasets is different.

\subsection{Greedy policy search for ensembles}
\label{subsec:ens}

\begin{figure}[t]
\centering
\includegraphics[width=\columnwidth]{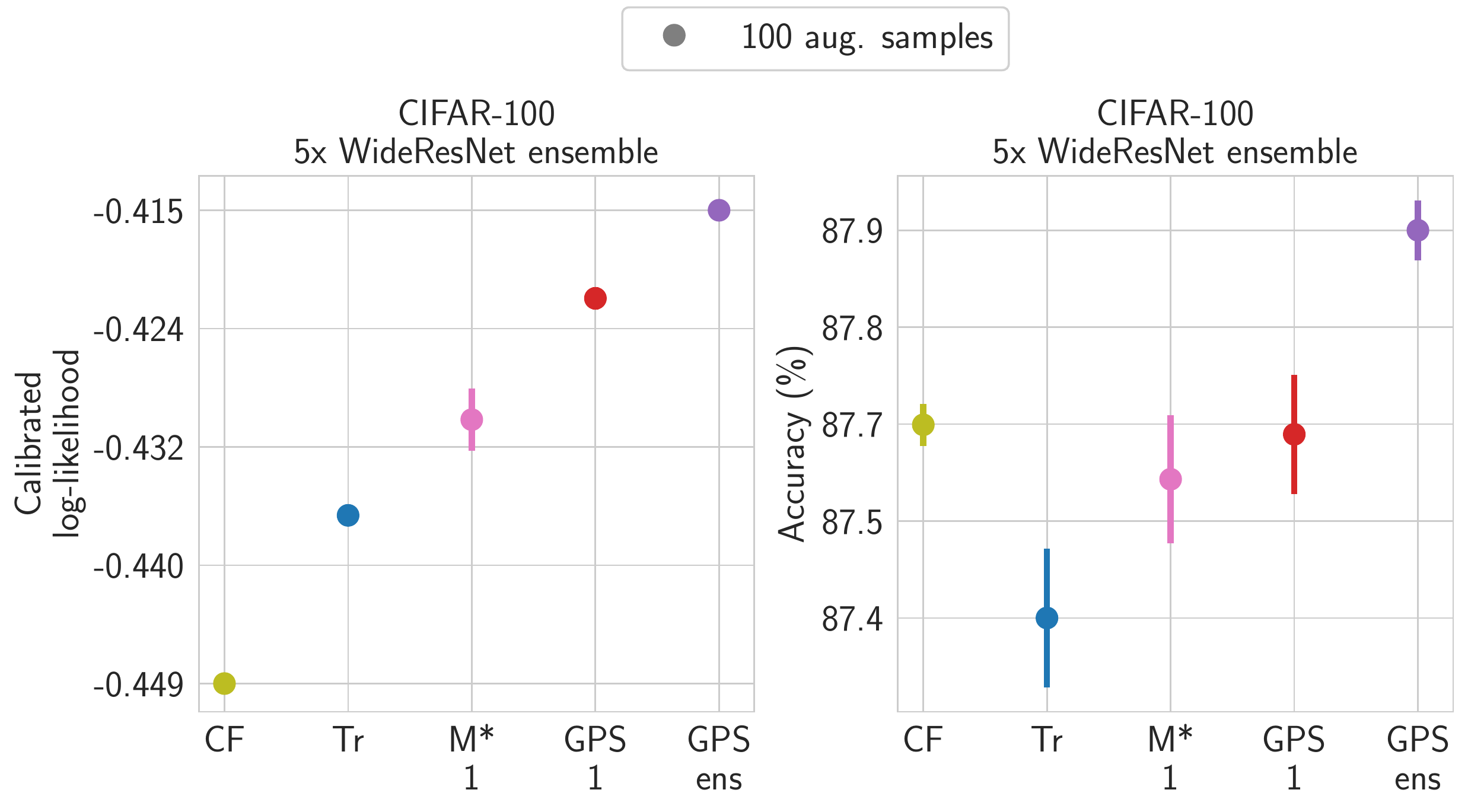}
\caption{ 
Greedy policy search improves the predictive performance of ensembles.
CC: central crop.
CF: random crops and horizontal flips. 
Tr: augmentation used for training (modified RandAugment with $\mathrm{M}=45$).
\mbox{``$\mathrm{M}^*$ 1''}: modified RandAugment with $\mathrm{M}^*=35$ found by grid search for a single model.
``GPS 1'': GPS is applied to a single model, and the ensemble is evaluated using the resulting policy.
``GPS ens'': GPS is applied to the whole ensemble.
{The results have been averaged over five runs of TTA.}
}
\label{fig:ens}
\end{figure}

Deep ensemble \citep{lakshminarayanan2017simple} is a simple yet powerful technique that achieves state-of-the-art results in in-domain and out-of-domain uncertainty estimation \citep{ovadia2019can,Ashukha2020Pitfalls}.
\cite{Ashukha2020Pitfalls} have shown that deep ensembles can be improved for free using test-time augmentation.
We show that deep ensembles can be improved even further by using a learnable test-time augmentation policy.

We use an ensemble of five WideResNet28x10 models, trained independently using the same training procedure as we used for training individual models (modified \mbox{RandAugment} training with $N=3$ and $M=45$).

There are several ways to apply GPS to an ensemble.
The simplest way is to perform GPS for a single model, and then evaluate the whole ensemble using that policy.
Another way is to perform GPS for the ensemble directly, using the same sub-policy for every member of the ensemble.
Other modifications can include searching for a separate policy for each member of the ensemble.
We test the first two options (denoted ``GPS single'' and ``GPS ensemble'' respectively), and leave other possible directions for future research.

The results are presented in Figure~\ref{fig:ens}.
They are consistent with the findings in previous sections.
Even a grid search for the optimal magnitude in test-time RandAugment is enough to significantly outperform random crops and flips. GPS improves the performance even further.
Transferring the policy from a single model to an ensemble (``GPS single'') performs worse than applying GPS to the whole ensemble directly, however, both variants of GPS outperform other baselines.
\begin{figure}[t]
\centering
\includegraphics[width=0.9\columnwidth]{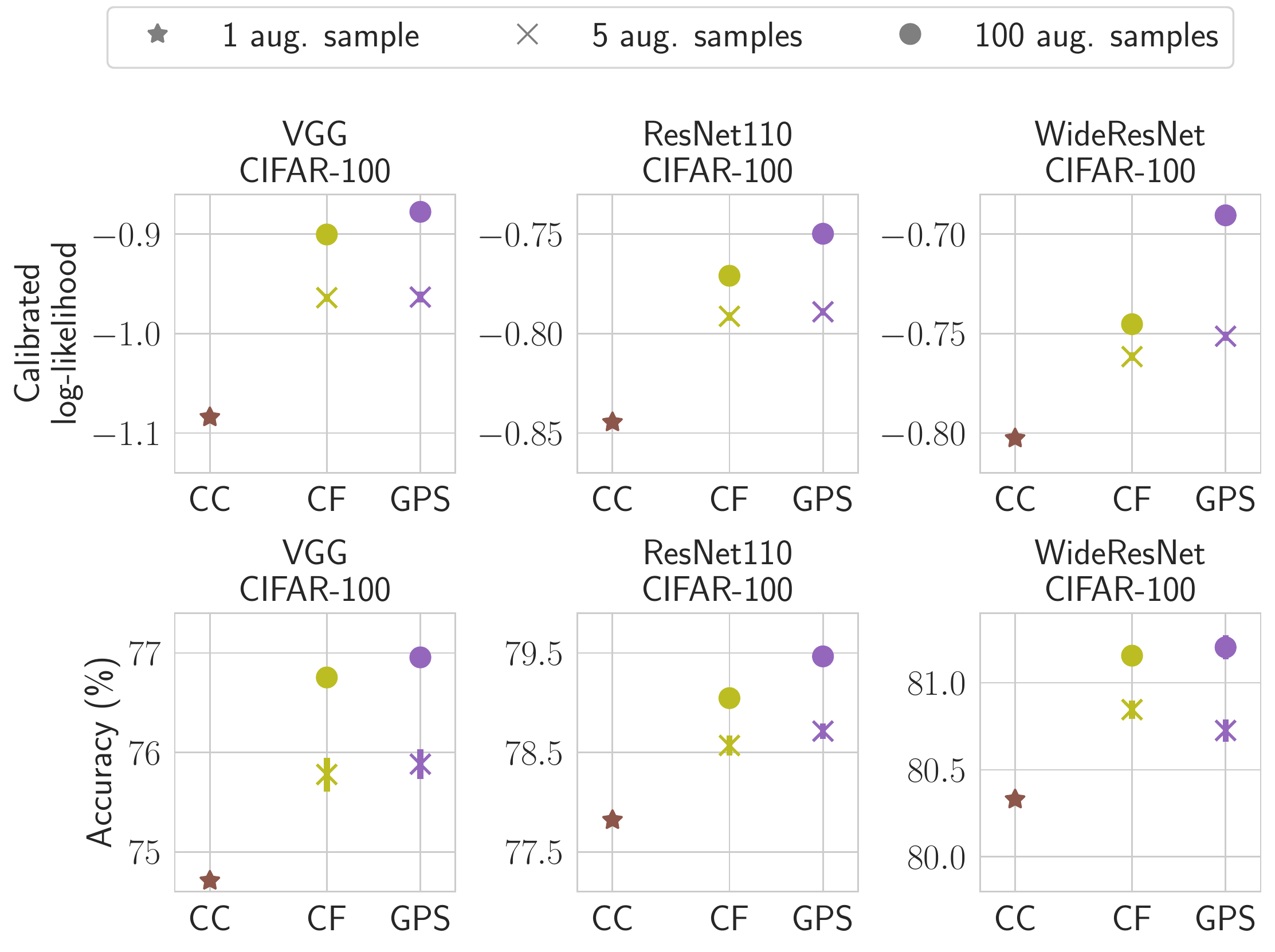}
\caption{
Greedy policy search (GPS) for models trained with vanilla augmentation (random crops and flips) still outperforms vanilla test-time  augmentation.
CC: central crop.
CF: random crops and horizontal flips.
GPS: greedy policy search.
{The results for CIFAR-100 have been averaged over five runs of TTA.}
}
\label{fig:clean-cifar100}
\end{figure}


The combination of ensembling methods and test-time augmentation usually provides meaningful benefits to predictive performance \citep{Ashukha2020Pitfalls}.
Because of this, we expect these results to also hold for other ensembling methods that are more efficient in terms of training time than deep ensembles.

\subsection{{Greedy policy search for models trained with vanilla augmentation}}
\label{subsec:clean}
While we mainly tested GPS for models trained with advanced data augmentation methods like RandAugment, it can be applied to any image classification model.
To further study the breadth of applicability of GPS, we apply it for models trained with standard (vanilla) data augmentation.
While the learned augmentation policy is less diverse than the policy learned for models trained with RandAugment (see Figure~\ref{fig:ra_augs_cifars}), GPS still manages to find a policy that significantly outperforms standard crops and flips on CIFAR-100 (see Figures~\ref{fig:clean-cifar100}~and~\ref{fig:in-domain-cifar10} for the comparison).
Even though the models learned with standard data augmentation are less robust to Rand\-Augment perturbations (see Figure~\ref{fig:mgrid}), they can benefit from some of the transformations.
The magnitude of the transformations is almost twice as low as compared to the policies for RandAugment models, and the identity transform is chosen much more often (see Figure~\ref{fig:transform_hist}).

\section{CONCLUSION}
We have designed a simple yet powerful greedy policy search method for test-time augmentation and tested it in a broad empirical evaluation.
To highlight the general idea that switching to learnable test-time augmentation strategy is beneficial, we aimed to keep the policy search simple rather than to tweak it for maximum performance.
Our findings can be summarized as follows:

\begin{itemize}
\item{We show that the learned test-time augmentation policies consistently provide superior predictive performance and uncertainty estimates compared to existing approaches to test-time augmentation.
{We report a significant improvement for both clean (in-domain) data and corrupted data (under domain shift).}}

\item{We find that the calibrated log-likelihood is a superior objective for learning test-time augmentation strategies as compared to LL or accuracy.
This finding may have important implications in adjacent fields such as meta-learning and neural architecture search, where the target (meta-)objective is often chosen to be either accuracy or plain validation log-likelihood with no calibration.
}

\item{We show the policies obtained with our method to be transferable between different architectures.
This means that transferring policies found for small architectures to large architectures is a viable strategy if computational resources are limited.}
\end{itemize}

There are many promising directions for future research on trainable test-time data augmentation.
{One potential area of improvement is in the design of dynamic object-dependent TTA policies as opposed to static object-independent policies, used in this paper.
Intuitively, this might be especially helpful under domain shift, as an object-dependent policy has a potential to alleviate it.}

\section*{Acknowledgements}

Dmitry Vetrov and Dmitry Molchanov were supported by the Russian Science Foundation grant
no. \mbox{19-71-30020}. This research was supported in part through computational resources of HPC
facilities at NRU HSE.

\bibliography{references}  
\bibliographystyle{icml2016}

\appendix

\section{Experimental details}
\label{sec:training_details}
{We train all our CIFAR models using a modified version of RandAugment.
The original implementation of RandAugment\footnote{\url{https://github.com/tensorflow/tpu/blob/master/models/official/efficientnet/autoaugment.py}}
mismatches the procedure described in the original paper.
The parameters of some transformations are ill-defined in the range of magnitudes used by RandAugment.
Some transformations also seem to become less severe as the magnitude parameter increases.
In our modification we've addressed these issues.
The full list of transformations with their parameters, along with their ranges used during training (corresponding to $M=45$) is presented below:}

\begin{tabular}{lll}
\toprule
    Transformation & \begin{tabular}[c]{@{}c@{}}Parameters\\ ($\widetilde{M}\sim\mathcal{U}\left[0, M\right]$)\end{tabular} &  \begin{tabular}[c]{@{}c@{}}Range for\\$M=45$\end{tabular} \\ 
    \midrule
    
    \texttt{Identity} & w/o parameters \\
    \texttt{ShearX} & $v=\widetilde{M}/60$ & $0\dots 75\%$ \\
    \texttt{ShearY} & $v=\widetilde{M}/60$ & $0\dots 75\%$ \\
    \texttt{TranslateX} & $v=0.015\cdot \widetilde{M}$ & $0\dots 67.5\%$ \\
    \texttt{TranslateY} & $v=0.015\cdot \widetilde{M}$ & $0\dots 67.5\%$ \\
    \texttt{Rotate} & $v=\frac43\widetilde{M}$ & $0\dots 60^{\circ}$ \\
    \texttt{Autocontrast} & $v=\frac13\widetilde{M}$ & $0\dots 15$ \\
    \texttt{Solarize} & $v=256-\frac{64}{15}\cdot \widetilde{M}$ & $256\dots 64$ \\
    \texttt{SolarizeAdd} & $v=256-\frac{64}{15}\cdot \widetilde{M}$ & $256\dots 64$ \\
    \texttt{Posterize} & $v=\max(0, 8-0.2\cdot \widetilde{M})$ & $8\dots 0$  \\
    \texttt{Contrast} & $v=\frac{2}{75}\cdot \widetilde{M}$ & $0\dots 1.2$  \\
    \texttt{Brightness} & $v=\frac{2}{75}\cdot \widetilde{M}$ & $0\dots 1.2$ \\
    \texttt{Color} & $v=0.03\cdot \widetilde{M}$ & $0\dots 1.35$ \\
    \texttt{Sharpness} & $v=0.03\cdot \widetilde{M}$ & $0\dots 1.35$ \\
    \texttt{Cutout} & $v=\widetilde{M}/60$ & $0\dots 75\%$ \\
    \bottomrule
\end{tabular}

{We apply the contrast, brightness, color and sharpness transformations in the following way: \texttt{PIL.ImageEnhance.OP.enhance(1+s*v)}, where \texttt{OP} stands for the name of the transformation, and $s$ is a random sign ($s\sim\mathcal{U}\{-1, +1\}$).}

{We also use mirrored background instead of the black background used by RandAugment to preserve the statistics of the augmented images and mitigate the potential problems caused by applying batch normalizaton to varying data.
Another, alternative way to mitigate the stated problems is to use different batch normalization statistics for different kinds of transformations, as proposed by \cite{xie2019adversarial}.}

\begin{figure*}[t]
\centering
\includegraphics[width=0.9\textwidth]{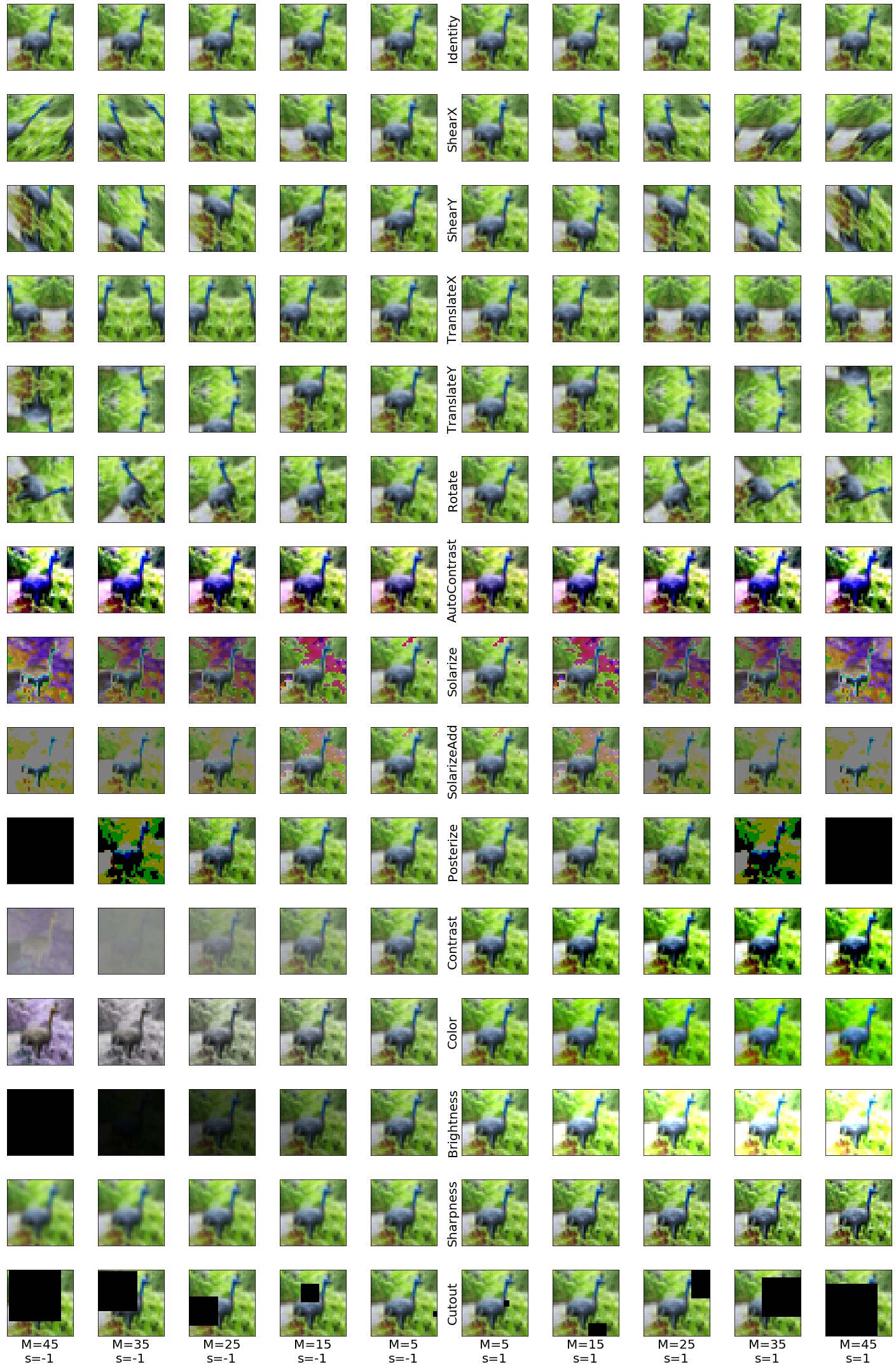}
\caption{Samples from the transformations used by our modification of RandAugment.}
\label{fig:ra}
\end{figure*}

{The samples of different transformations for different magnitudes are presented in Figure~\ref{fig:ra}.
The magnitude of each transformation $\widetilde{M}$ during training is resampled from the uniform distribution $\widetilde{M}\sim\mathcal{U}\left[0, M\right]$.
Each application of RandAugment is followed by a random crop and flip. }

{With a crude grid search over the magnitudes $M$, we find that the optimal magnitude for training is $M=45$ for all the considered models on CIFAR-10 and CIFAR-100.
However, we find that when networks are trained with augmentations that are so strong, longer training is essential.
We train all models for 2000 epochs of SGD with momentum and a step decay schedule, dividing the learning rate in half 12 times during training.
All models used the initial learning rate of $0.1$ and the momentum of $0.9$.
VGG and PreResNet110 used the weight decay of $0.0003$, and WideResNet28x10 used the weight decay of $0.0005$.}

\textbf{Generation of the pool of sub-policies~} 
The policy pool for CIFAR models was created with the same version of RandAugment as we used for training.
We sample $500$ sub-policies from RandAugment with $N=3$ and $M=45$, $500$ sub-policies with $N=3$ and $M=20$, $100$ sub-policies with $N=3$ and $M=0$\footnote{Sub-policies with zero magnitude mostly consist of identity transformations, with an occasional autocontrast being present in some sub-policies.}, and use the identity transformation as the final sub-policy, resulting in a pool of $1101$ sub-policies.
The magnitude of each transformation in each sub-policy is sampled from the uniform distribution $\mathcal{U}\left[0, M\right]$ and fixed.
Each application of a GPS sub-policy to a CIFAR image is followed with a random crop and flip.

The first transformation in each ImageNet sub-policy was fixed to the standard scale-crop-flip transformation with a learnable magnitude of scaling.
We did not count the scale-crop-flip transformation in the length of sub-policies $N.\ $
In addition to the transformations used for CIFAR datasets, the ImageNet list of transformations also contains \texttt{Invert} and \texttt{Equalize}.
For ImageNet models we sample $300$ sub-policies with $N=2$ and $M=45$, $300$ sub-policies with $N=2$ and $M=20$, $100$ sub-policies with $N=3$ and $M=10$, $100$ sub-policies with $N=1$ and $M=45$, and $100$ sub-policies 
consisting only of scale-crop-flip transformations.

\newpage
\clearpage
\onecolumn
\section{Additional experimental results}

\begin{figure}[h!]
\centering
\includegraphics[width=0.65\columnwidth]{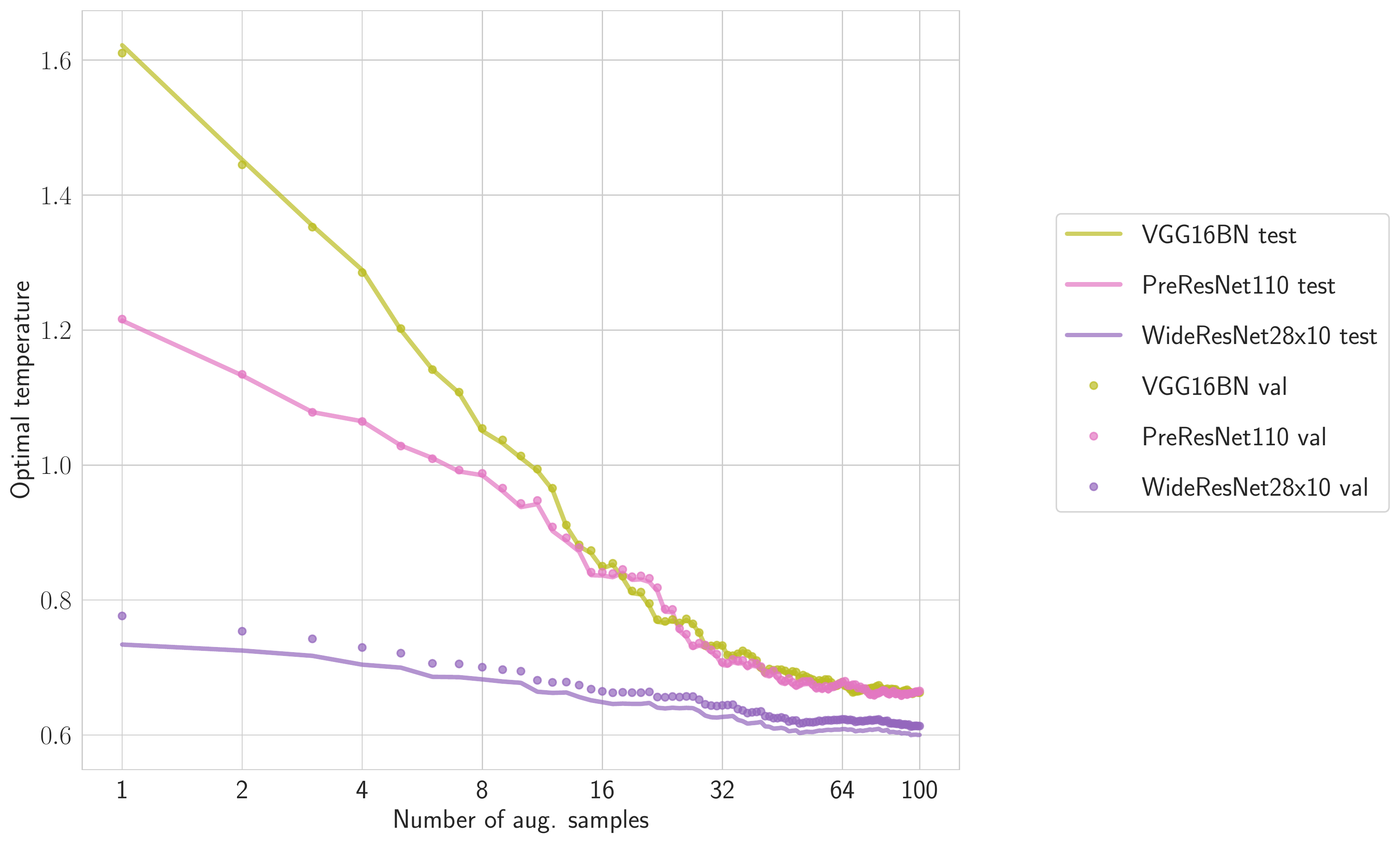}
\caption{
Optimal temperature for different numbers of samples drawn from the GPS policy.
The optimal temperature decreases with the number of samples for all models.
The optimal temperature found on the validation set matches the temperature found by test-time cross-validation.
}
\label{fig:temp}
\end{figure}

\begin{figure}[h!]
\centering
\includegraphics[width=0.85\textwidth]{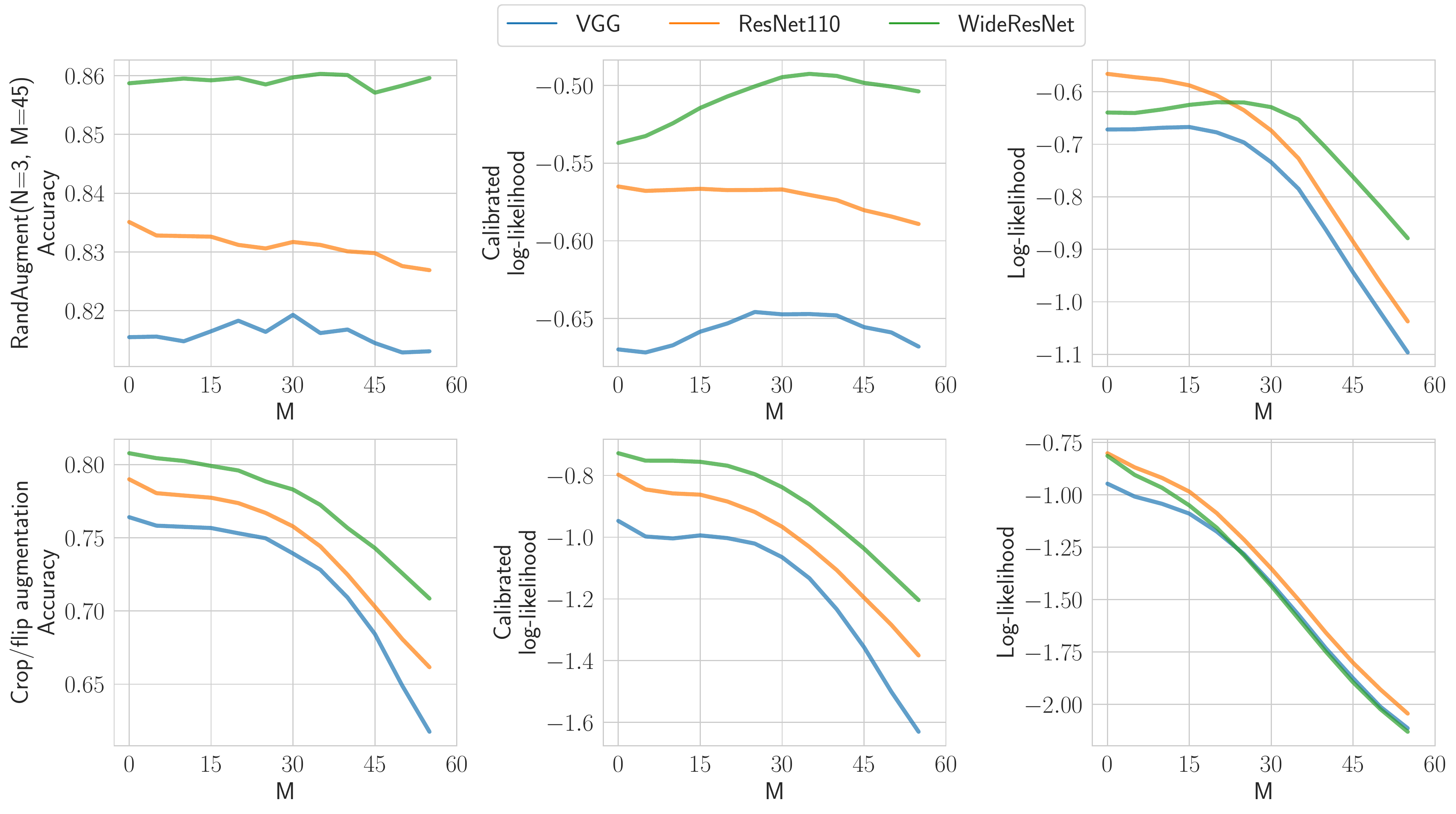}
\caption{
The predictive performance of a 100-sample test-time augmentation with a RandAugment policy with $N=3$ and with different magnitudes $M$ from $0$ to $55$.
The results are presented for different models, trained on CIFAR-100 with RandAugment (top) and standard data augmentation (bottom).
This plot highlights several properties of test-time augmentation.
First, we see that the optimal magnitude $M$ for test-time augmentation is different for different models and is not necessarily equal to the magnitude used during training.
Second, we see that the larger the magnitude, the more decalibrated the models become (the gap between the log-likelihood and the calibrated log-likelihood increases with $M$).
This means that the log-likelihood is failing to evaluate the performance of TTA fairly.
Third, the models trained with standard crops and flips have a much lower base performance (at $M=0$) and quickly degrade as the magnitude increases.
It means that plain magnitude grid search is not enough to obtain a TTA policy that would outperform random crops and flips.
Interestingly, greedy policy search still manages to find a policy that outperforms this baseline (see Section~\ref{subsec:clean} for details).
}
\label{fig:mgrid}
\end{figure}

\newpage
\clearpage

\newpage
\clearpage
\begin{figure*}[t]
\centering
\includegraphics[width=0.42\textwidth]{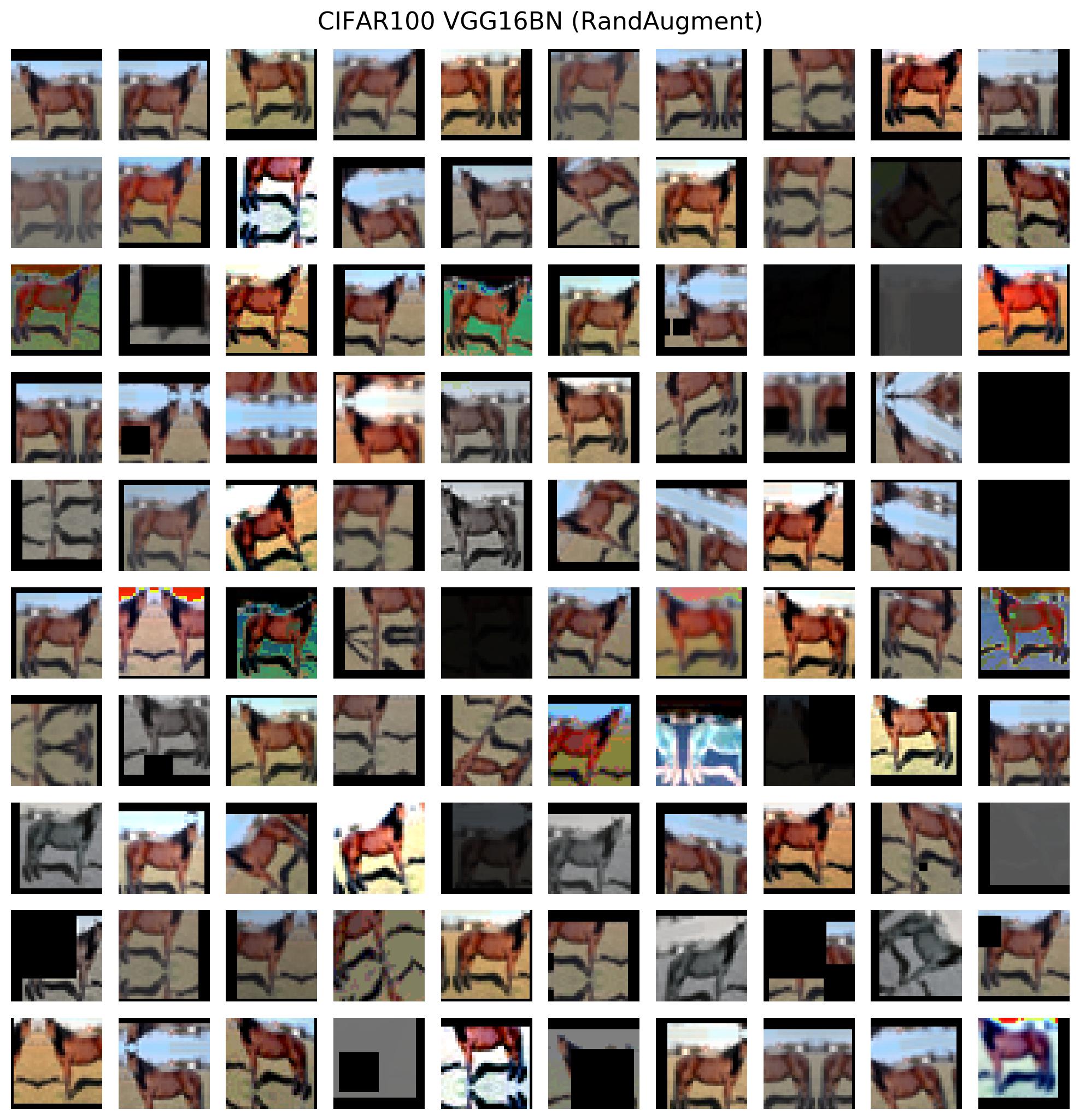}
~
\includegraphics[width=0.42\textwidth]{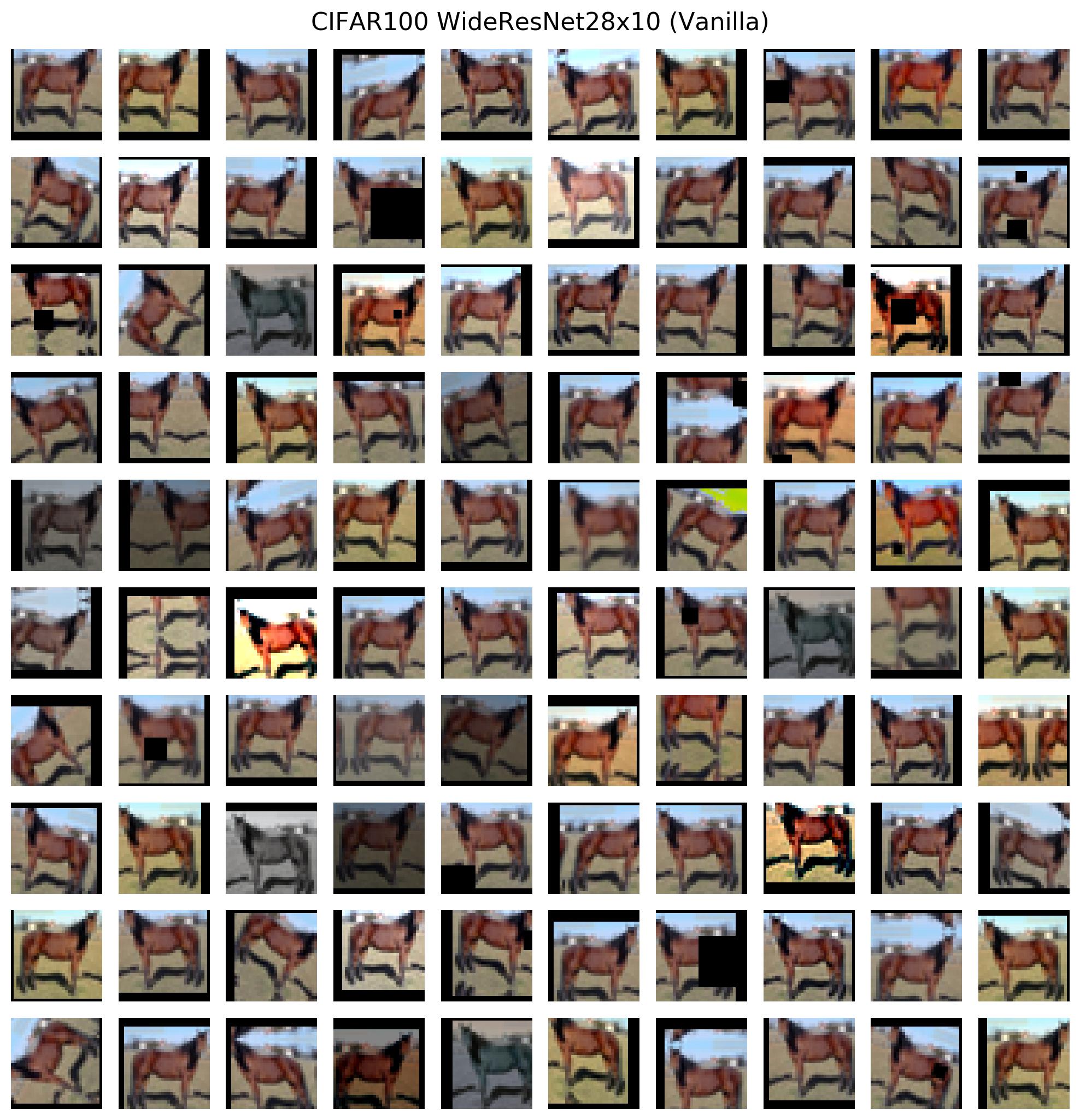}\\
\includegraphics[width=0.42\textwidth]{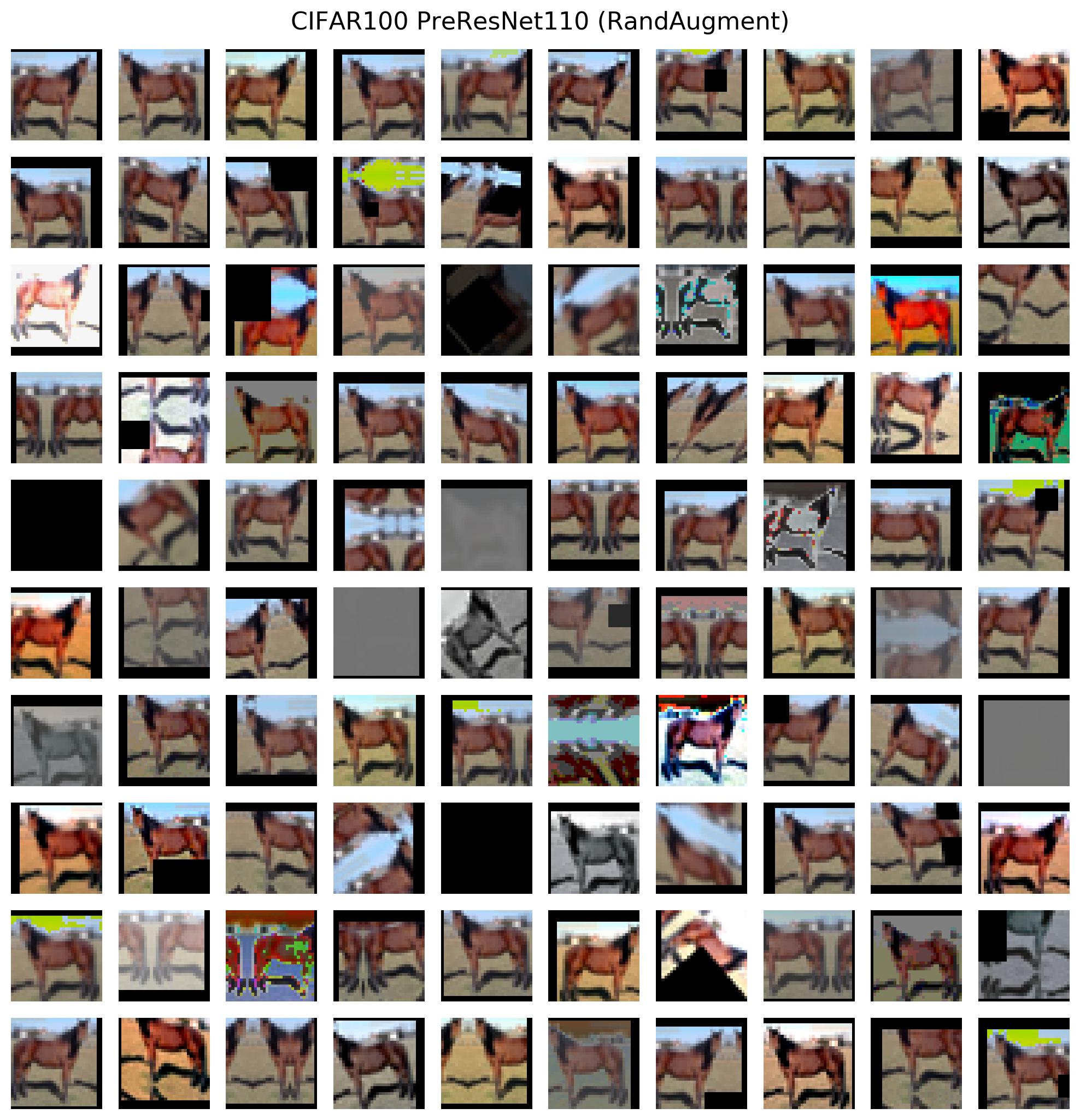}
~
\includegraphics[width=0.42\textwidth]{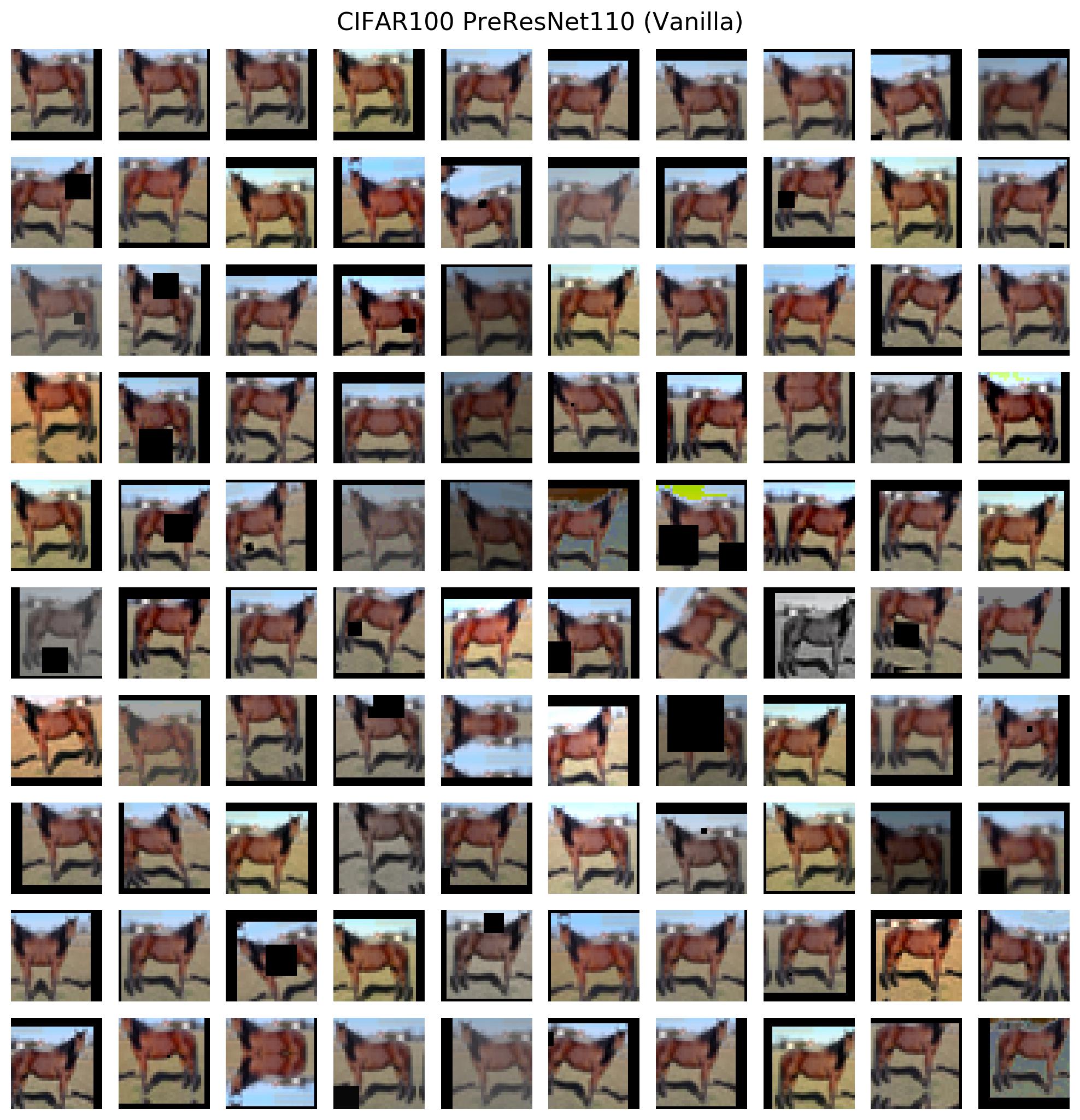}\\
\includegraphics[width=0.42\textwidth]{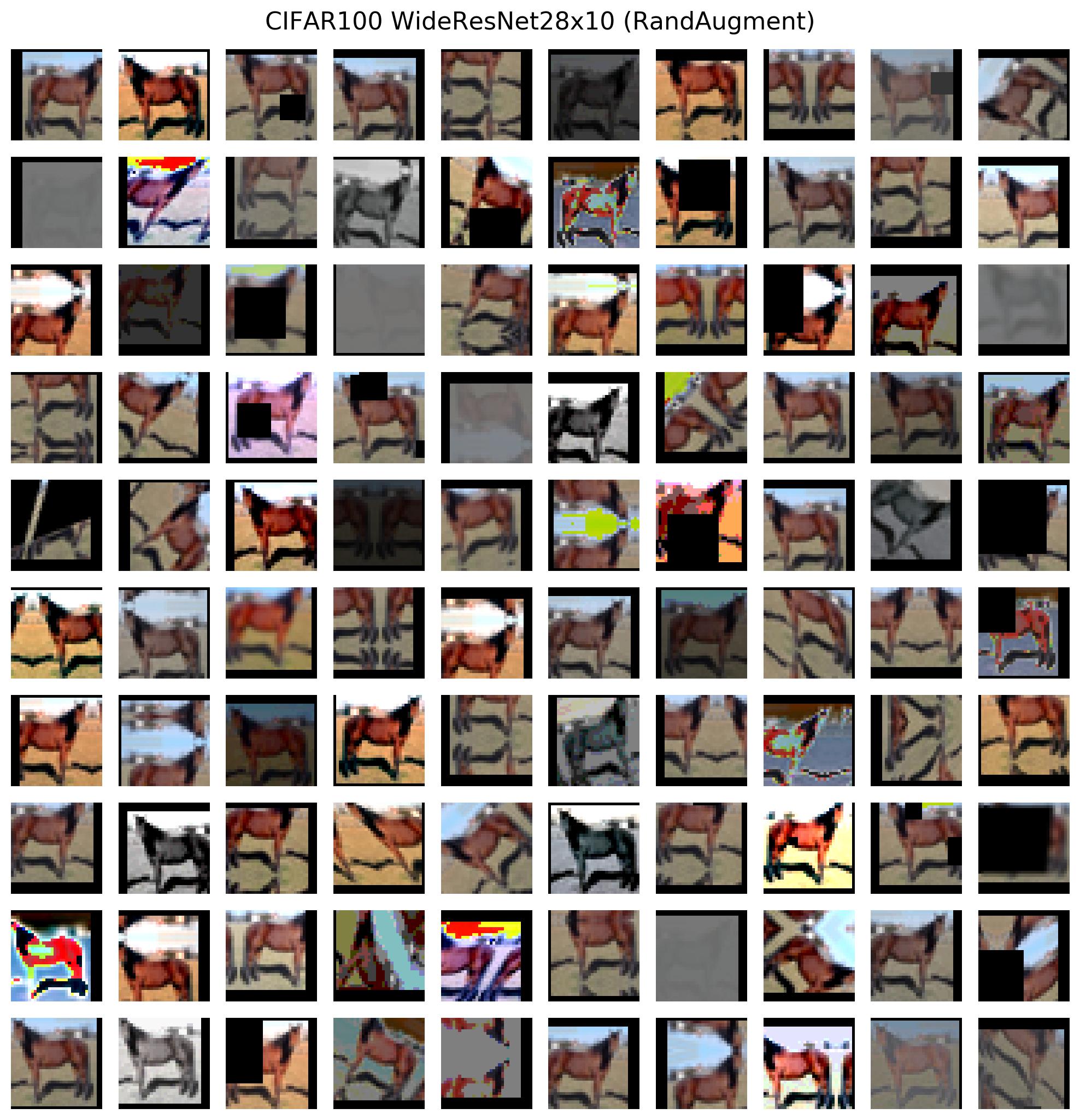}
~
\includegraphics[width=0.42\textwidth]{imgs/CIFAR100_WideResNet28x10_Vanilla_policy_npz.jpg}
\caption{Policies learned by GPS for different models on CIFAR-100, models in the left column were trained with RandAugment, and models on the right were trained with vanilla crop-flip augmentation.
Policies consist of 100 sub-policies, 1 sample from each is shown.
Training initial model with RandAugment allows GPS to choose more diverse sub-policies.
}
\label{fig:ra_augs_cifars}
\end{figure*}

\begin{figure*}[t]
\centering
\includegraphics[width=0.32\textwidth]{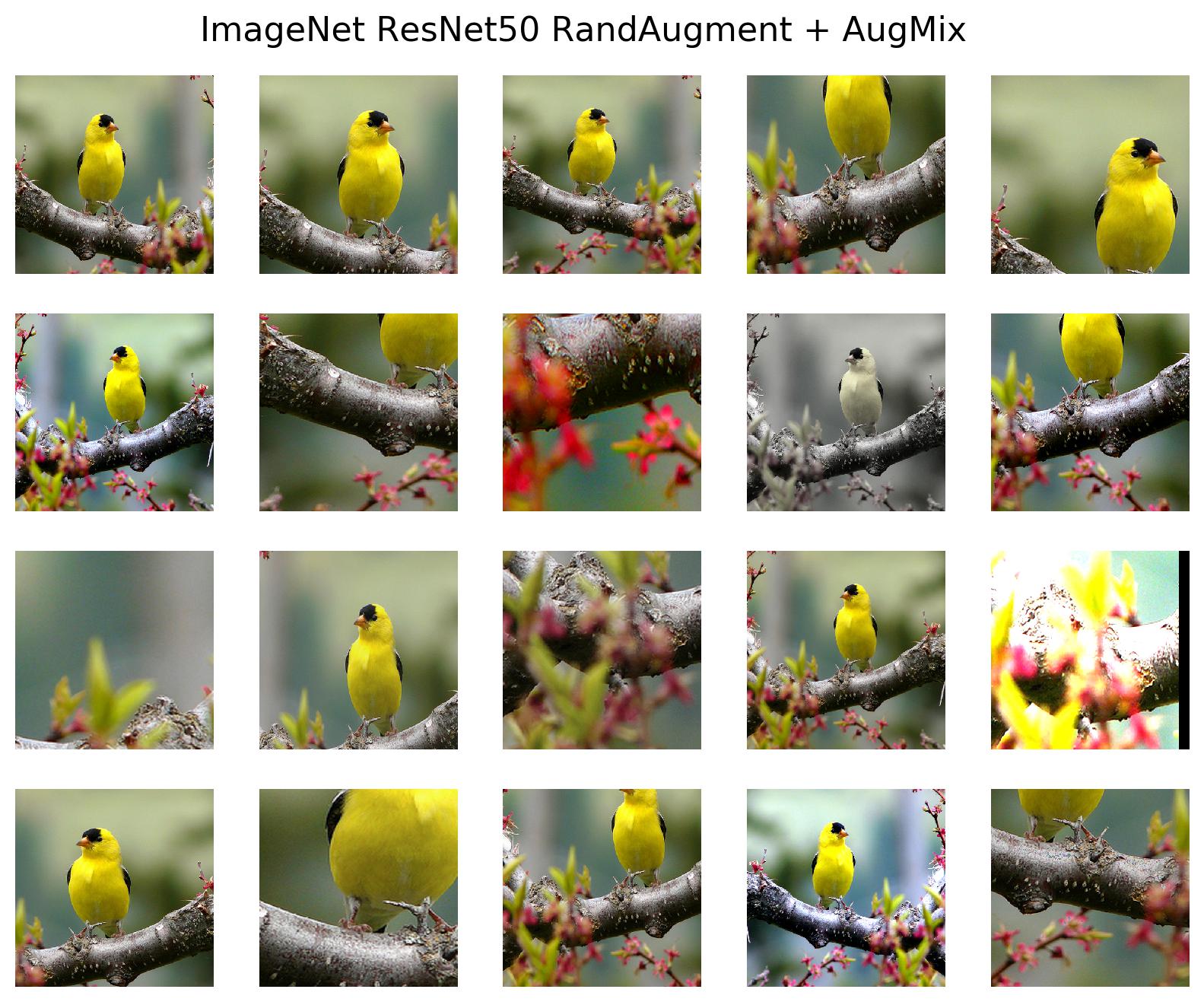}
~
\includegraphics[width=0.32\textwidth]{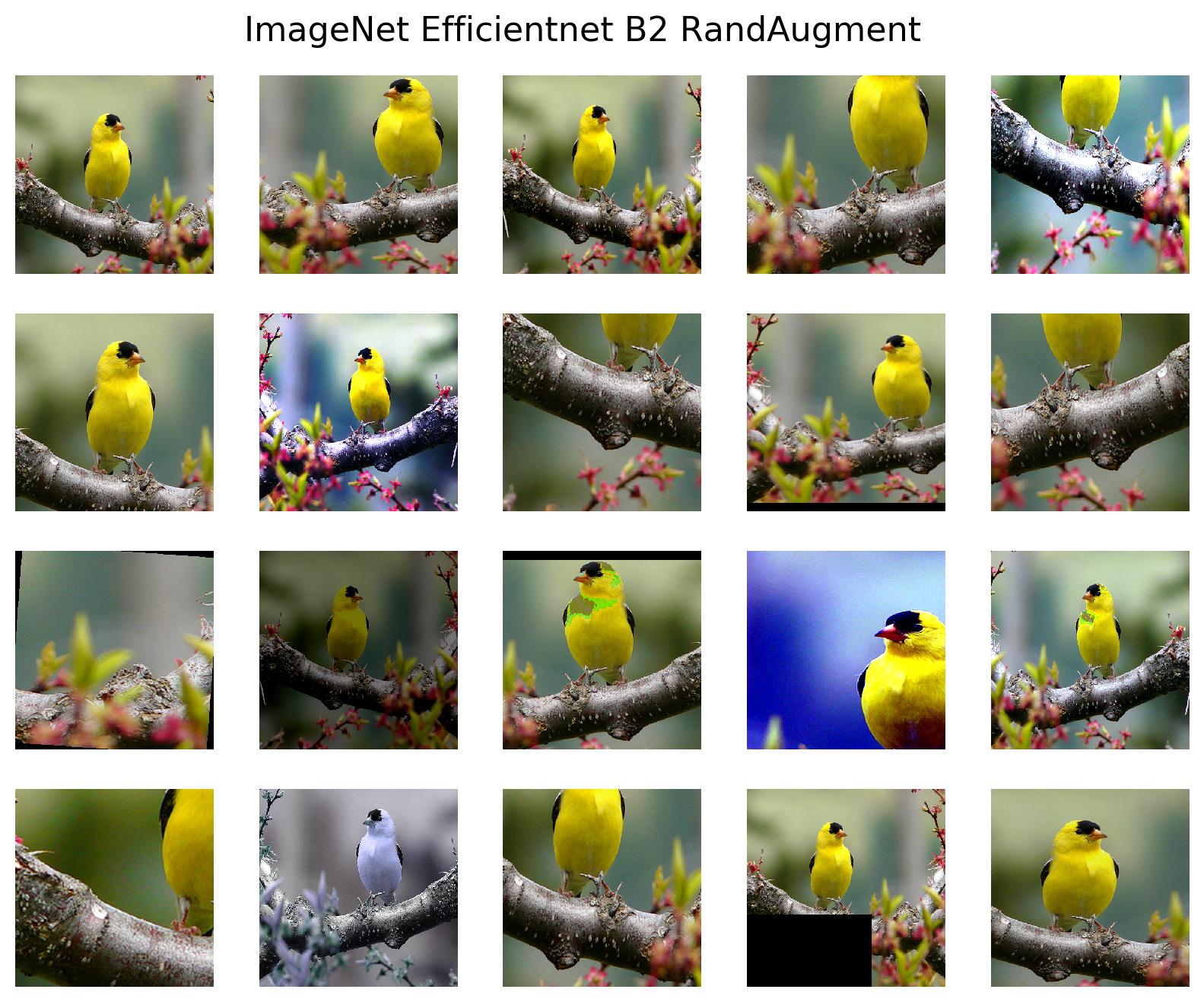}
~
\includegraphics[width=0.32\textwidth]{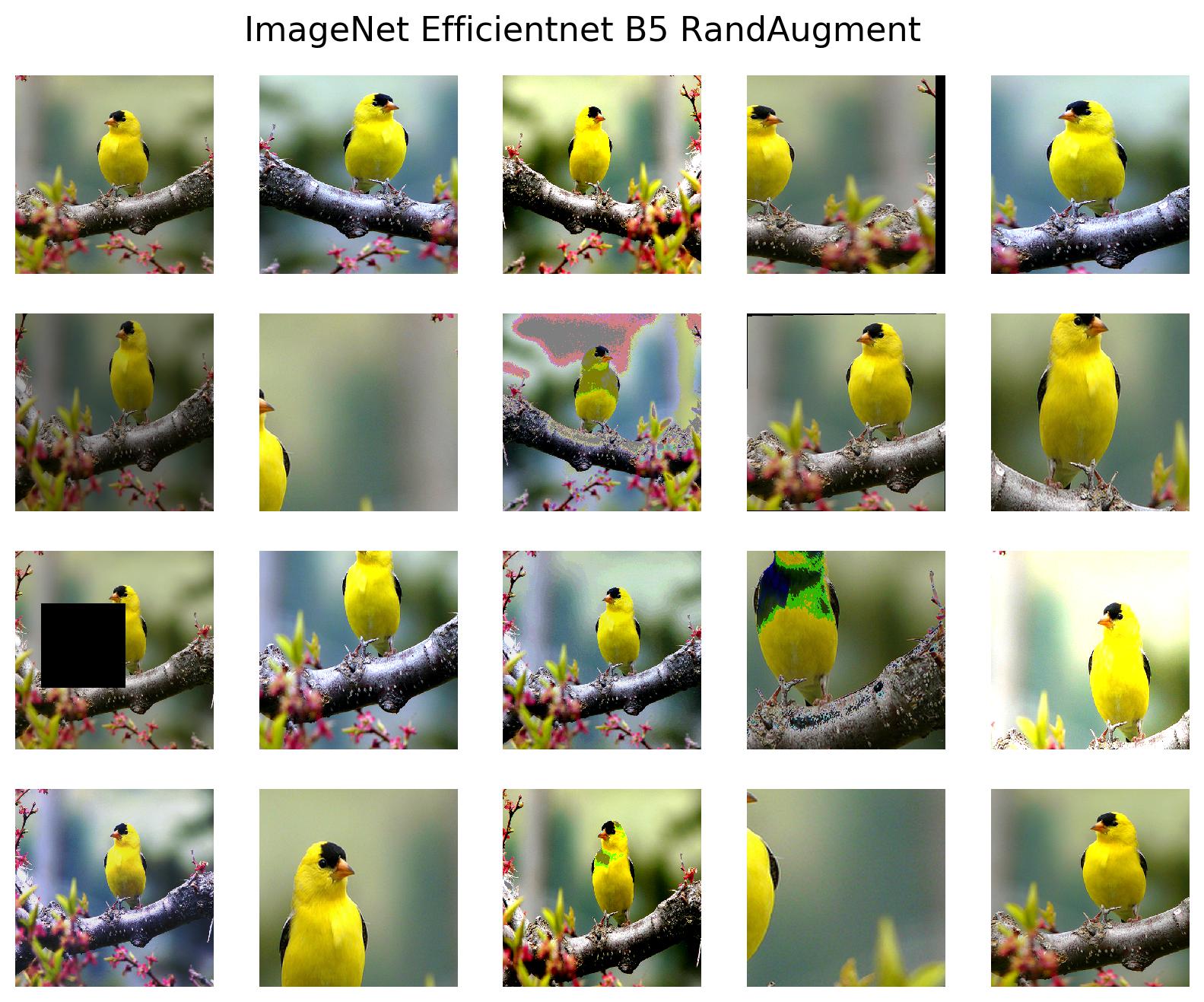}
\caption{Policies learned by GPS for different models on ImageNet.
Each policy consists of 20 sub-policies, 1 sample from each sub-policy is shown.
}
\label{fig:ra_augs_imagenet}
\end{figure*}

\begin{figure*}[t]
\includegraphics[width=0.495\textwidth]{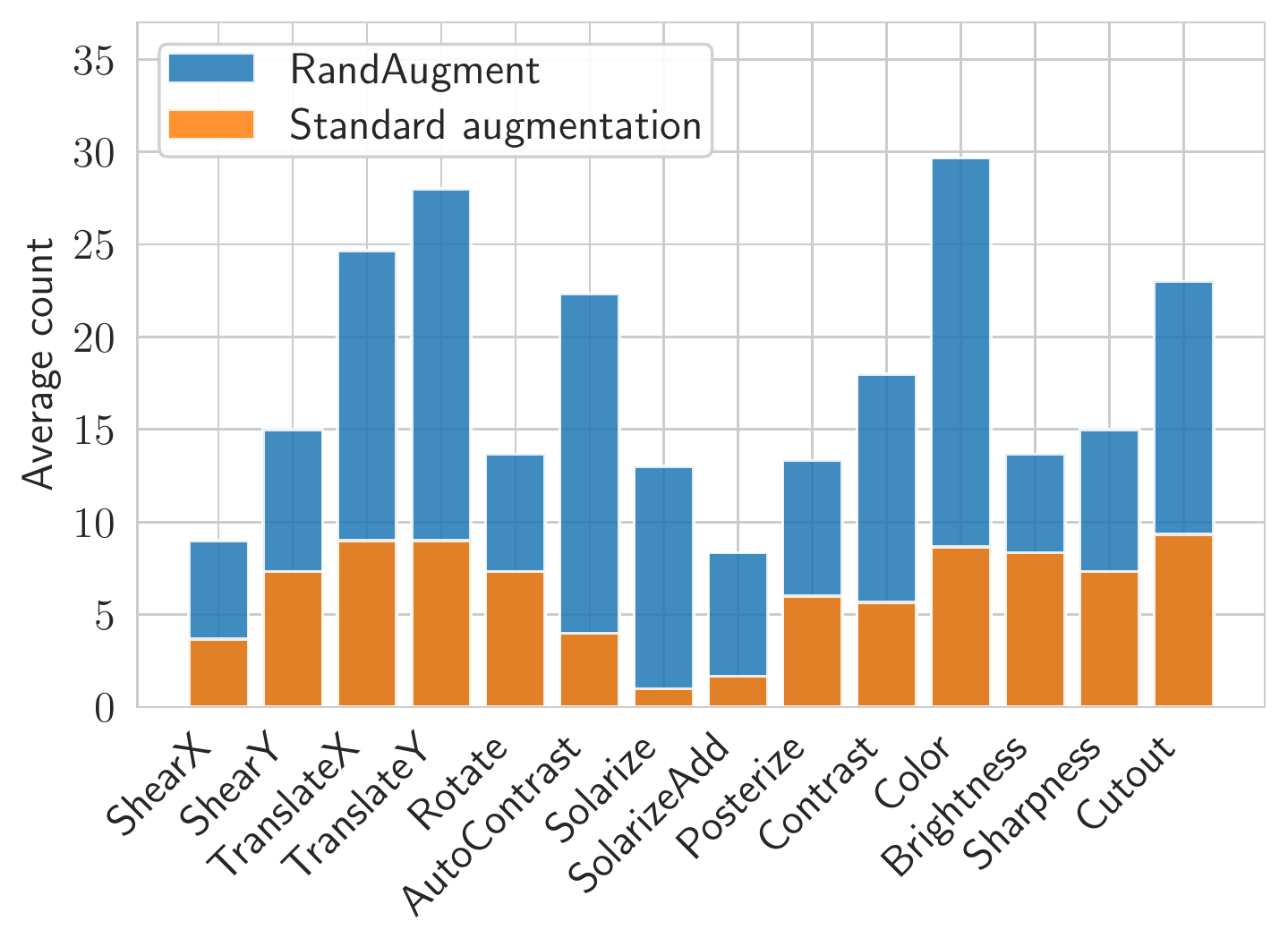}
\includegraphics[width=0.495\textwidth]{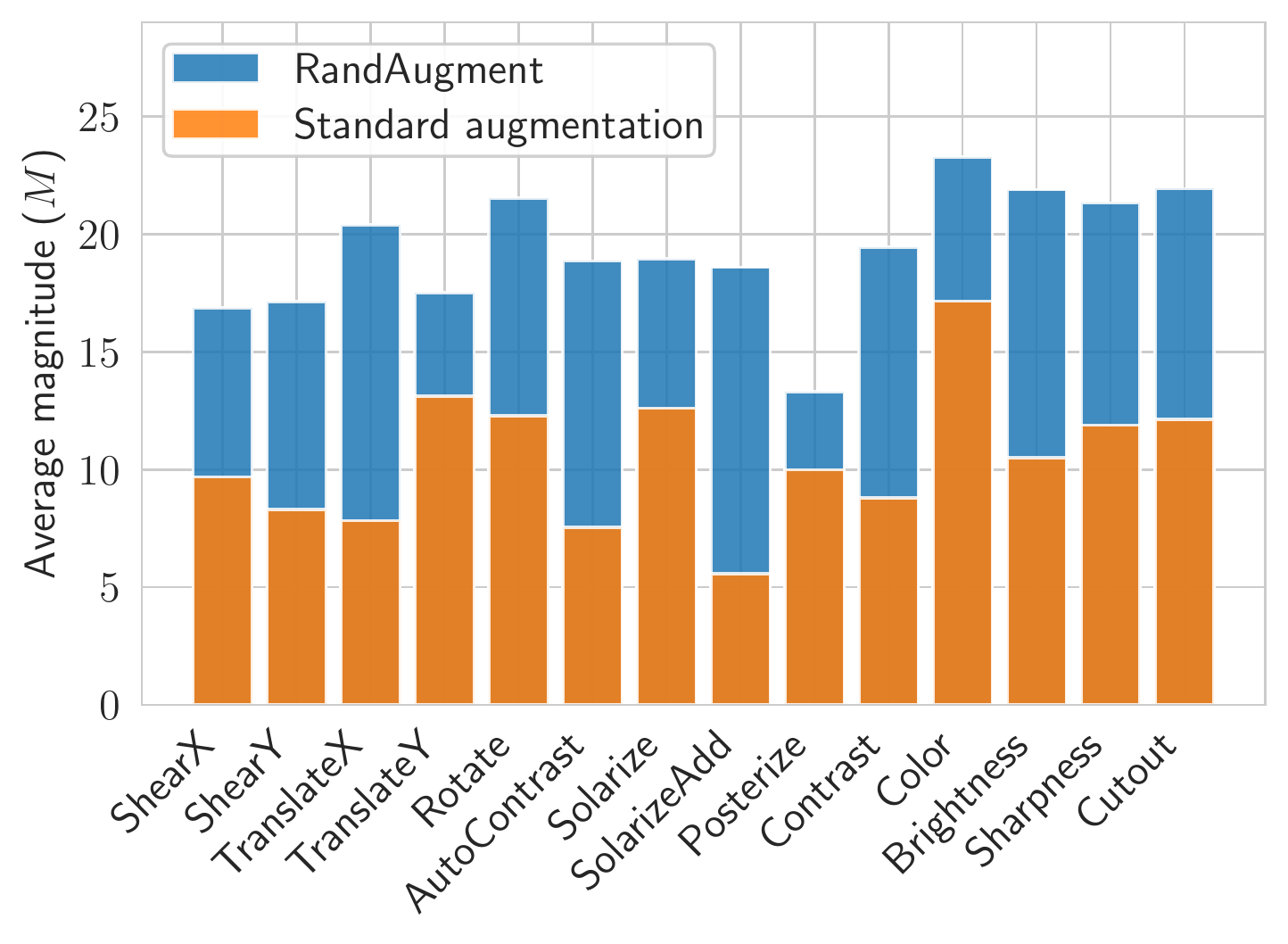}
\caption{
Counts and average magnitudes for each non-trivial transformation in learned GPS policies. The values are compared between models trained with standard data augmentation consisting of horizontal flips and crops and models trained with RandAugment.
Values are averaged over VGG, ResNet110 and WideResNet architectures for CIFAR-100 dataset.
Since the number of sub-policies in a policy is fixed, policies for models trained with standard augmentation contain a significantly larger number of identity transformations (the overall count of non-identity transformations is lower).
}
\label{fig:transform_hist}
\end{figure*}

\begin{figure*}[t]
\includegraphics[width=\textwidth]{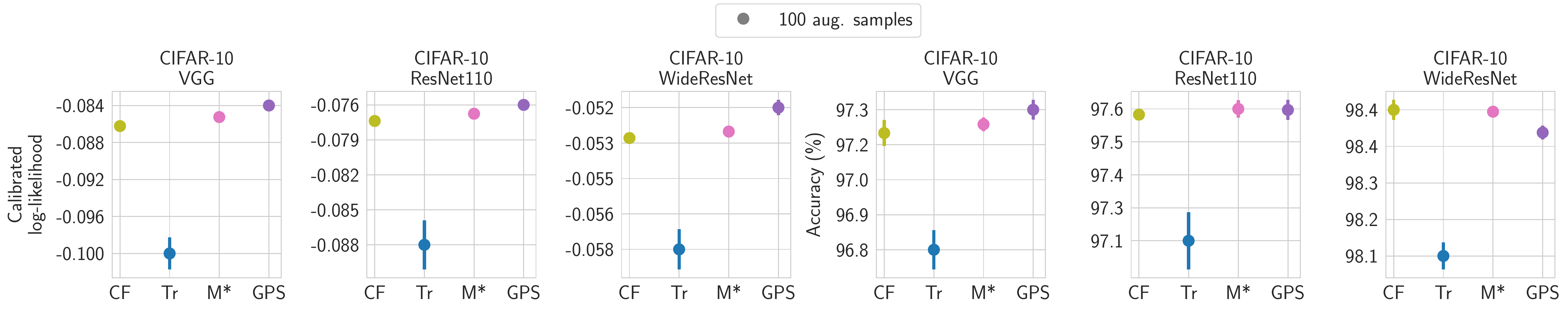}
\caption{
Performance of various test-time augmentation strategies on the clean test set of CIFAR-10 dataset.
Greedy policy search (GPS) outperforms all other methods in terms of calibrated log-likelihood and achieves a comparable accuracy.
}
\label{fig:in-domain-cifar10}
\end{figure*}

\clearpage
\newpage
\begin{table*}
\centering
\resizebox{1.\textwidth}{!}{
\begin{tabular}{llllllllllllllllllll}
\toprule

 &&\multicolumn{9}{c}{\footnotesize Calibrated log-likelihood} &\multicolumn{9}{c}{\footnotesize Accuracy (\%)} \\
 \cmidrule(l){3-11} \cmidrule(l){12-20}
Dataset &    Model        &    \multicolumn{1}{c}{CC}  &    \multicolumn{1}{c}{5C}   &   \multicolumn{1}{c}{10C}   &  \multicolumn{3}{c}{CF} &  \multicolumn{3}{c}{GPS} &   \multicolumn{1}{c}{CC}  & \multicolumn{1}{c}{5C}&    \multicolumn{1}{c}{10C}  &  \multicolumn{3}{c}{CF} &  \multicolumn{3}{c}{GPS} \\
\cmidrule(l){3-3}\cmidrule(l){4-4}\cmidrule(l){5-5}\cmidrule(l){6-8}\cmidrule(l){9-11}\cmidrule(l){12-12}\cmidrule(l){13-13}\cmidrule(l){14-14}\cmidrule(l){15-17}\cmidrule(l){18-20}
 &  &    & &  &  \multicolumn{1}{c}{\scriptsize{5}} & \multicolumn{1}{c}{\scriptsize{10}} & \multicolumn{1}{c}{\scriptsize{20}} &  \multicolumn{1}{c}{\scriptsize{5}} & \multicolumn{1}{c}{\scriptsize{10}} & \multicolumn{1}{c}{\scriptsize{20}} &     &  &  & \multicolumn{1}{c}{\scriptsize{5}} & \multicolumn{1}{c}{\scriptsize{10}} & \multicolumn{1}{c}{\scriptsize{20}} & \multicolumn{1}{c}{\scriptsize{5}} & \multicolumn{1}{c}{\scriptsize{10}} & \multicolumn{1}{c}{\scriptsize{20}} \\
\cmidrule(l){1-1} \cmidrule(l){2-2} \cmidrule(l){3-3} \cmidrule(l){4-4} \cmidrule(l){5-5} \cmidrule(l){6-6} \cmidrule(l){7-7} \cmidrule(l){8-8} \cmidrule(l){9-9} \cmidrule(l){10-10} \cmidrule(l){11-11} \cmidrule(l){12-12} \cmidrule(l){13-13} \cmidrule(l){14-14} \cmidrule(l){15-15} \cmidrule(l){16-16} \cmidrule(l){17-17} \cmidrule(l){18-18} \cmidrule(l){19-19} \cmidrule(l){20-20}
  &         ResNet50 &  -0.859 &  -0.799 &  -0.791 &  -0.815 &  -0.779 &  -0.762 &  -0.785 &  -0.766 &  -0.755 &  79.21 &  80.00 &  80.07 &  79.82 &   80.32 &   80.63 &  80.42 &   80.69 &   80.93 \\
 ImageNet &  EfficientNet B2 &  -0.793 &  -0.729 &  -0.727 &  -0.754 &  -0.715 &  -0.696 &  -0.717 &  -0.695 &  -0.685 &  80.43 &  81.50 &  81.65 &  81.22 &   81.85 &   82.04 &  81.68 &   82.09 &   82.23 \\
  &  EfficientNet B5 &  -0.643 &  -0.614 &  -0.612 &  -0.628 &  -0.603 &  -0.590 &  -0.600 &  -0.586 &  -0.577 &  83.78 &  84.26 &  84.28 &  84.08 &   84.33 &   84.54 &  84.20 &   84.44 &   84.62 \\
\bottomrule
\end{tabular}
}
\caption{Calibrated log-likelihood and accuracy of all test-time augmentation methods on ImageNet: central crop (CC), 5/10-crop evaluation (four corner crops, one center crop for 5C; five crops with horizontal flip on/off for 10C), a policy consisting of random crops and flips (CF), greedy policy search (GPS).}
\label{table:acc-nll-ImageNet}
\end{table*}

\begin{table*}
\centering
\resizebox{1.\textwidth}{!}{
\begin{tabular}{llllllllllllllllllllllllllll}
\toprule
 &&\multicolumn{12}{c}{\footnotesize Calibrated log-likelihood} &\multicolumn{12}{c}{\footnotesize Accuracy (\%)} \\
\cmidrule(l){3-15} \cmidrule(l){16-28}
Dataset &            Model & \multicolumn{1}{c}{CC} & \multicolumn{3}{c}{CF} & \multicolumn{3}{c}{MTrain} & \multicolumn{3}{c}{MGrid} & \multicolumn{3}{c}{GPS} & \multicolumn{1}{c}{CC} & 
\multicolumn{3}{c}{CF}& \multicolumn{3}{c}{MTrain} & \multicolumn{3}{c}{MGrid}& \multicolumn{3}{c}{GPS} \\
\cmidrule(l){3-3}\cmidrule(l){4-6}\cmidrule(l){7-9}\cmidrule(l){10-12}\cmidrule(l){13-15}\cmidrule(l){16-16}\cmidrule(l){17-19}\cmidrule(l){20-22}\cmidrule(l){23-25}\cmidrule(l){26-28}
 &  & &  \multicolumn{1}{c}{\scriptsize{5}}& \multicolumn{1}{c}{\scriptsize{10}} & \multicolumn{1}{c}{\scriptsize{100}} &\multicolumn{1}{c}{\scriptsize{5}} & \multicolumn{1}{c}{\scriptsize{10}} & \multicolumn{1}{c}{\scriptsize{100}} &\multicolumn{1}{c}{\scriptsize{5}} & \multicolumn{1}{c}{\scriptsize{10}} & \multicolumn{1}{c}{\scriptsize{100}} & \multicolumn{1}{c}{\scriptsize{5}}& \multicolumn{1}{c}{\scriptsize{10}} & \multicolumn{1}{c}{\scriptsize{100}} &  & \multicolumn{1}{c}{\scriptsize{5}} & \multicolumn{1}{c}{\scriptsize{10}} & \multicolumn{1}{c}{\scriptsize{100}} &\multicolumn{1}{c}{\scriptsize{5}} & \multicolumn{1}{c}{\scriptsize{10}} & \multicolumn{1}{c}{\scriptsize{100}} & \multicolumn{1}{c}{\scriptsize{5}} & \multicolumn{1}{c}{\scriptsize{10}} & \multicolumn{1}{c}{\scriptsize{100}} & \multicolumn{1}{c}{\scriptsize{5}} & \multicolumn{1}{c}{\scriptsize{10}} & \multicolumn{1}{c}{\scriptsize{100}} \\
\cmidrule(l){1-1} \cmidrule(l){2-2} \cmidrule(l){3-3} \cmidrule(l){4-4} \cmidrule(l){5-5} \cmidrule(l){6-6} \cmidrule(l){7-7} \cmidrule(l){8-8} \cmidrule(l){9-9} \cmidrule(l){10-10} \cmidrule(l){11-11} \cmidrule(l){12-12} \cmidrule(l){13-13} \cmidrule(l){14-14} \cmidrule(l){15-15} \cmidrule(l){16-16} \cmidrule(l){17-17} \cmidrule(l){18-18} \cmidrule(l){19-19} \cmidrule(l){20-20} \cmidrule(l){21-21} \cmidrule(l){22-22} \cmidrule(l){23-23} \cmidrule(l){24-24} \cmidrule(l){25-25} \cmidrule(l){26-26} \cmidrule(l){27-27} \cmidrule(l){28-28}
  &     PreResNet110 &    -0.090 &  -0.082 &  -0.080 &   -0.077 &     -0.132 &      -0.107 &       -0.088 &    -0.081 &     -0.079 &      -0.077 &  -0.081 &   -0.080 &    -0.076 &     97.17 &  97.50 &   97.54 &    97.61 &      96.12 &       96.66 &        97.13 &     97.58 &      97.60 &       97.64 &   97.52 &    97.53 &     97.63 \\
  CIFAR10 &          VGG &    -0.110 &  -0.093 &  -0.089 &   -0.086 &     -0.158 &      -0.122 &       -0.100 &    -0.093 &     -0.089 &      -0.085 &  -0.094 &   -0.091 &    -0.084 &     96.86 &  97.09 &   97.19 &    97.23 &      95.48 &       96.36 &        96.75 &     97.11 &      97.21 &       97.27 &   97.11 &    97.25 &     97.33 \\
  &  WideResNet28x10 &    -0.066 &  -0.056 &  -0.054 &   -0.053 &     -0.090 &      -0.071 &       -0.058 &    -0.056 &     -0.054 &      -0.053 &  -0.056 &   -0.054 &    -0.052 &     98.04 &  98.36 &   98.39 &    98.45 &      97.49 &       97.90 &        98.12 &     98.32 &      98.37 &       98.44 &   98.27 &    98.35 &     98.40 \\
\cmidrule(r){1-1}\cmidrule{2-2}\cmidrule(lr){3-15}\cmidrule(lr){16-28}
  &     PreResNet110 &    -0.626 &  -0.588 &  -0.579 &   -0.573 &     -0.736 &      -0.645 &       -0.569 &    -0.585 &     -0.577 &      -0.569 &  -0.588 &   -0.576 &    -0.552 &     81.67 &  82.80 &   83.05 &    83.21 &      79.89 &       81.82 &        83.21 &     82.91 &      83.15 &       83.28 &   82.94 &    83.27 &     83.49 \\
 CIFAR100 &          VGG &    -0.852 &  -0.736 &  -0.714 &   -0.689 &     -0.909 &      -0.771 &       -0.645 &    -0.767 &     -0.709 &      -0.643 &  -0.732 &   -0.690 &    -0.624 &     78.07 &  80.51 &   80.89 &    81.30 &      77.21 &       79.68 &        81.72 &     79.92 &      80.99 &       81.81 &   80.64 &    81.31 &     82.11 \\
  &  WideResNet28x10 &    -0.636 &  -0.571 &  -0.562 &   -0.555 &     -0.669 &      -0.573 &       -0.493 &    -0.591 &     -0.541 &      -0.491 &  -0.553 &   -0.519 &    -0.479 &     84.06 &  85.21 &   85.36 &    85.53 &      82.46 &       84.64 &        86.12 &     84.15 &      85.20 &       86.07 &   85.22 &    85.89 &     86.38 \\
\bottomrule
\end{tabular}
}
\caption{Calibrated log-likelihood and accuracy of all test-time augmentation methods on CIFAR-10/100: central crop (CC), a policy consisting of random crops and flips (CF), RandAugment with magnitude learned for training (MTrain), RandAugment with magnitude learned for test (MGrid), greedy policy search (GPS).}
\label{table:acc-nll-cifars}
\end{table*}

\begin{table*}
\centering
\resizebox{0.5\textwidth}{!}{
\begin{tabular}{llcccccccccc}
\toprule
&& \multicolumn{10}{c}{Mean unnormalized corruption error (muCE)}\\
\cmidrule(l){3-12}
Dataset &       Model & \multicolumn{1}{c}{CC} &  \multicolumn{3}{c}{CF} & \multicolumn{3}{c}{Mgrid} & \multicolumn{3}{c}{GPS} \\
\cmidrule(l){3-3}\cmidrule(l){4-6}\cmidrule(l){7-9} \cmidrule(l){10-12}   
      &        &  &  \multicolumn{1}{c}{\scriptsize{5}} & \multicolumn{1}{c}{\scriptsize{10}} & \multicolumn{1}{c}{\scriptsize{100}} & \multicolumn{1}{c}{\scriptsize{5}} & \multicolumn{1}{c}{\scriptsize{10}} & \multicolumn{1}{c}{\scriptsize{100}} & \multicolumn{1}{c}{\scriptsize{5}} & \multicolumn{1}{c}{\scriptsize{10}} & \multicolumn{1}{c}{\scriptsize{100}} \\
\cmidrule(l){1-1} \cmidrule(l){2-2} \cmidrule(l){3-3} \cmidrule(l){4-4} \cmidrule(l){5-5} \cmidrule(l){6-6} \cmidrule(l){7-7} \cmidrule(l){8-8} \cmidrule(l){9-9} \cmidrule(l){10-10} \cmidrule(l){11-11} \cmidrule(l){12-12}
 &   ResNet110 &    10.930 &  10.218 &  10.056 &    9.922 &    10.158 &      9.972 &       9.820 &  10.117 &    9.594 &     9.439 \\
  CIFAR10-C &         VGG &    11.652 &  10.559 &  10.313 &   10.068 &    10.457 &     10.188 &       9.956 &  10.344 &   10.010 &     9.716 \\
 &  WideResNet &     7.957 &   7.362 &   7.233 &    7.112 &     7.342 &      7.202 &       7.078 &   7.098 &    6.947 &     6.715 \\
\cmidrule(r){1-1}\cmidrule{2-2} \cmidrule(lr){3-12}
 &   ResNet110 &    35.307 &  33.998 &  33.742 &   33.544 &    33.950 &     33.601 &      33.289 &  33.595 &   32.412 &    31.875 \\
 CIFAR100-C &         VGG &    36.804 &  34.021 &  33.507 &   32.971 &    34.728 &     32.711 &      30.882 &  33.123 &   32.141 &    30.848 \\
 &  WideResNet &    30.786 &  29.286 &  29.068 &   28.858 &    29.588 &     28.060 &      26.421 &  29.312 &   28.234 &    26.649 \\
\bottomrule
\end{tabular}

}
\caption{Mean unnormalized corruption error (muCE) of test-time augmentation methods on CIFAR-10-C/CIFAR-100-C: central crop (CC), a policy consisting of random crops and flips (CF), RandAugment with magnitude learned for test (MGrid), greedy policy search (GPS).}
\label{table:ce-cifarsC}
\end{table*}

\begin{table*}
\centering
\resizebox{0.4\textwidth}{!}{
\begin{tabular}{lllllllll}
\toprule
&& \multicolumn{7}{c}{Mean corruption error (mCE)}\\
\cmidrule(l){3-9}
     Dataset &            Model &  \multicolumn{1}{c}{CC}    &  \multicolumn{3}{c}{CF} & \multicolumn{3}{c}{GPS} \\
\cmidrule(l){3-3}\cmidrule(l){4-6}\cmidrule(l){7-9} 
      &             &      & \multicolumn{1}{c}{\scriptsize{5}} & \multicolumn{1}{c}{\scriptsize{10}} & \multicolumn{1}{c}{\scriptsize{20}} & \multicolumn{1}{c}{\scriptsize{5}} & \multicolumn{1}{c}{\scriptsize{10}} & \multicolumn{1}{c}{\scriptsize{20}} \\
\cmidrule(l){1-1} \cmidrule(l){2-2} \cmidrule(l){3-3} \cmidrule(l){4-4} \cmidrule(l){5-5} \cmidrule(l){6-6} \cmidrule(l){7-7} \cmidrule(l){8-8} \cmidrule(l){9-9} 
  &         ResNet50 &  0.687 &  0.723 &   0.708 &   0.701 &  0.699 &   0.675 &   0.673 \\
 ImageNet-C &  EfficientNet B2 &  0.655 &  0.680 &   0.665 &   0.658 &  0.646 &   0.641 &   0.638 \\
  &  EfficientNet B5 &  0.559 &  0.609 &   0.594 &   0.586 &  0.548 &   0.544 &   0.538 \\
\bottomrule
\end{tabular}
}
\caption{Mean corruption error (mCE) of test-time augmentation methods on ImageNet-C: central crop (CC), a policy consisting of random crops and flips (CF), greedy policy search (GPS).}
\label{table:ce-ImageNetC}
\end{table*}

\end{document}